\documentclass{article}

\usepackage[preprint]{neurips_2025}

\usepackage[utf8]{inputenc}
\usepackage[T1]{fontenc}
\usepackage{amsmath,amssymb,amsfonts}
\usepackage{mathtools}
\usepackage{bm}
\usepackage{booktabs}
\usepackage{graphicx}
\usepackage{hyperref}
\usepackage{cleveref}
\usepackage{algorithm}
\usepackage{algpseudocode}   
\usepackage{amsthm}
\usepackage{subfig}
\usepackage{natbib}
\usepackage{xcolor}

\newcommand{\vtheta}{\bm{\theta}}
\newcommand{\vz}{\bm{z}}
\newcommand{\mH}{\bm{H}}

\newcommand{\logZ}{\log \mathcal{Z}}

\newcommand{\data}{\mathcal{D}}

\title{Automatic Laplace Collapsed Sampling: Scalable Marginalisation of Latent Parameters via Automatic Differentiation}

\author{
  Toby Lovick \\
  Institute of Astronomy, University of Cambridge \\
  \And
  David Yallup \\
  Kavli Institute for Cosmology, University of Cambridge \\
  \And
  Will Handley \\
  Institute of Astronomy \& Kavli Institute for Cosmology, University of Cambridge \\
}

\begin{document}

\maketitle

\begin{abstract}
We present Automatic Laplace Collapsed Sampling (ALCS), a general framework for marginalising latent parameters in Bayesian models using automatic differentiation, which we combine with nested sampling to explore the hyperparameter space in a robust and efficient manner.
At each nested sampling likelihood evaluation, ALCS collapses the high-dimensional latent variables $\vz$ to a scalar contribution via maximum a posteriori (MAP) optimisation and a Laplace approximation, both computed using autodiff.
This reduces the effective dimension from $d_\theta + d_z$ to just $d_\theta$, making Bayesian evidence computation tractable for high-dimensional settings without hand-derived gradients or Hessians, and with minimal model-specific engineering. The MAP optimisation and Hessian evaluation are parallelised across live points on GPU-hardware, making the method practical at scale.
We also show that automatic differentiation enables local approximations beyond Laplace to parametric families such as the Student-$t$, which improves evidence estimates for heavy-tailed latents.
We validate ALCS on a suite of benchmarks spanning hierarchical, time-series, and discrete-likelihood models and establish where the Gaussian approximation holds. This enables a post-hoc ESS diagnostic that localises failures across hyperparameter space without expensive joint sampling.
\end{abstract}

\section{Introduction}
\label{sec:intro}

Bayesian model comparison via the marginal likelihood (evidence) $\mathcal{Z} = \int \mathcal{L}(\data \mid \vtheta)\,\pi(\vtheta)\,d\vtheta$ is central to scientific inference.
It provides a principled mechanism for comparing models of different complexity, automatically penalising over-parameterisation through the Occam factor without requiring cross-validation~\citep{mackay1992}.
Nested sampling~\citep{Skilling2006} has become a standard algorithm for computing $\mathcal{Z}$ directly in the physical sciences~\citep{Ashton2022,Buchner2023}, maintaining a live particle cloud that compresses the prior to the posterior while accumulating contributions to the evidence integral.

The fundamental limitation of nested sampling is its scaling with parameter dimension $d$: to maintain fixed precision in $\log \mathcal{Z}$, the number of required likelihood evaluations scales roughly as $\mathcal{O}(d^3)$, arising from the need to scale both the number of live points and the number of inner sampler steps with $d$~\citep{Yallup2026,Hu2024,kroupa2025resonances}.
The dominant paradigm for scaling to high dimensions, Hamiltonian Monte Carlo (HMC)~\citep{Neal2011} and its variants~\citep{Hoffman2014}, achieves better dimension scaling~\citep{beskos2010optimaltuninghybridmontecarlo} but does not directly estimate $\mathcal{Z}$, requiring additional approximations such as bridge sampling~\citep{gelman_simulating_1998} or harmonic mean estimators~\citep{McEwen2021} that can introduce systematic errors.

Many of the problems that most demand scalable evidence computation share a deeper common structure: a set of \emph{parameters of interest} $\vtheta$ governing the population or physical model, and a high-dimensional set of \emph{latent variables} $\vz$ representing per-object measurement noise, calibration offsets, or discretised field values whose individual realisations are not scientifically interesting but whose uncertainty must be correctly propagated into the posterior on $\vtheta$.
This structure arises across a broad range of high-impact astrophysical pipelines: hierarchical supernova cosmology~\citep{BAYESN, brout2022}, galaxy population modelling~\citep{popcosmos}, gravitational-wave population inference, and exoplanet atmospheric retrievals all reduce to this $(\vtheta, \vz)$ factorisation.
Marginalising $\vz$ given $\vtheta$ is, in essence, automatic Gaussianised error propagation: it converts per-object measurement uncertainty into a $\vtheta$-dependent uncertainty on the population parameters, accounting for the fact that the sensitivity of the observables to measurement noise changes as a function of the model being evaluated.
This $\vtheta$-dependence is what current practice sacrifices for tractability: large-scale analyses such as Pantheon+~\citep{brout2022} marginalise per-supernova nuisances into a static covariance matrix at a single reference cosmology, so the propagated uncertainty does not reflect the true model sensitivity at each $\vtheta$ under test.

This paper introduces \emph{Automatic Laplace Collapsed Sampling} (ALCS), a general framework for marginalising latent parameters in Bayesian models, that combines the strengths of nested sampling for robustly exploring the parameters of interest in a model, with the speed of modern autodiff optimization methods for fast approximation of latent variables. 
ALCS collapses $\vz$ at each evaluation of the marginal likelihood using local gradient information: it optimises $\vz$ to its conditional MAP and computes the Gaussian evidence contribution via automatic differentiation, so that any outer sampler operates exclusively in the lower-dimensional $\vtheta$ space.

A key enabling insight comes from differentiable physical forward models such as $\partial$Lux~\citep{desdoigts2024}, which demonstrated that automatic differentiation makes exact second derivatives tractable at $d_z \sim 10^5$.
Wherever a forward model is differentiable, the full Hessian of the joint log-posterior is automatically accessible, which is the computational barrier that previously made Laplace marginalisation a bespoke, per-problem exercise.
For the evidence-computation experiments in this paper we use the BlackJAX nested sampling implementation~\citep{NSSyallup} as the outer loop, exploiting its composability with JAX primitives, but the collapsed marginal likelihood interface is sampler-agnostic. Within nested sampling, parallel live-point evaluation maps naturally to GPU batching~\citep{Yallup2026}, and the auto-grad requirements of ALCS are met wherever JAX-based forward models exist, as an increasing share of inference pipelines in cosmology and beyond are ported to GPUs.

We explore the efficacy of the approach on synthetic problems designed to probe the boundaries of the Gaussian assumption, standard high-dimensional Bayesian inference benchmarks, and a scalable application to modern survey supernova cosmology.
We motivate ALCS both as a general-purpose tool for high-dimensional sampling problems and by identifying high-impact astrophysical inference pipelines where this approach is well suited.
Specifically, we make the following contributions:
\begin{enumerate}
\item[\textit{(i)}] \textbf{ALCS framework.} We present ALCS as a general framework for Laplace marginalisation of latent parameters, requiring only a differentiable joint log-posterior. The collapsed marginal likelihood is composable with any outer sampler; we give explicit conditions under which the approach is appropriate and analyse its computational cost.
\item[\textit{(ii)}] \textbf{JAX implementation.} We provide a JAX implementation exploiting automatic differentiation for Hessians and \texttt{jax.vmap} for parallelism across blocks of latent variables, with prior whitening, warm-start preconditioning, and sparse Hessian support for structured latent spaces. The ``automatic'' in ALCS is deliberate: no hand-derived gradients, Hessians, or model-specific code are required, only a differentiable log-posterior.
\item[\textit{(iii)}] \textbf{Validation and scope.} We validate ALCS on synthetic benchmarks (a supernova cosmology model scaling to $d_z = 25{,}600$, a heavy-tailed Student-$t$ test, and a tanh funnel as a controlled failure case), six models from the inference gym, and we establish where the Gaussian assumption holds and where it fails via an importance-sampling ESS diagnostic. As a demonstration of scope, we show that the autodiff framework extends naturally beyond Gaussian approximation: matching the fourth derivative at the MAP to a Student-$t$ family improves evidence estimates for heavy-tailed latent posteriors.
\end{enumerate}

\section{Background}
\label{sec:background}

\subsection{Bayesian Model Comparison}
\label{subsec:model_comparison}

Given data $\data$ and a model $\mathcal{M}$ with parameters $\vtheta$, Bayesian inference targets the posterior distribution of our model's parameters updated by the data $p(\vtheta \mid \data) \propto \mathcal{L}(\data \mid \vtheta)\,\pi(\vtheta)$, where $\mathcal{L}$ is the likelihood and $\pi$ the prior.
The normalising constant,
\begin{equation}
\mathcal{Z} = \int \mathcal{L}(\data \mid \vtheta)\,\pi(\vtheta)\,d\vtheta,
\label{eq:evidence}
\end{equation}
is the \emph{Bayesian evidence} (or marginal likelihood).
When comparing two models $\mathcal{M}_1$ and $\mathcal{M}_2$, the Bayes factor $\mathcal{Z}_1 / \mathcal{Z}_2$ quantifies the relative support from the data~\citep{Jeffreys1961}. Calculation of this integral is a notoriously challenging task that has inspired the development of a variety of computational approaches~\citep{mlreview}. Usage of Bayes factors as a hypothesis testing framework is central to many statistical scientific applications, notably in astrophysics~\citep{Trotta_2008}, which makes extensive use both of the Bayes factor and nested sampling to perform the numerical integrals.
Unlike information criteria such as AIC or BIC, the evidence is a genuine integral over parameter space and automatically encodes Occam's razor: a model that assigns high likelihood only in a small region of prior space receives a large Occam penalty from the ratio of posterior to prior volume. Using the Jeffreys scale~\citep{Jeffreys1961}, strong comparisons can be made when evidences differ by $\Delta\log \mathcal{Z} \sim 2$, so a boundary of $\pm 1$ nats is used as the relative scale where the Laplace error would begin to affect decision making.

\subsection{The Laplace Approximation}
\label{subsec:laplace}

The Laplace approximation estimates the integral of an unnormalised density $p^*(\vz) = \mathcal{L}(\data \mid \vz)\,\pi(\vz)$ by fitting a Gaussian at its mode.
Expanding $\log p^*$ to second order around the MAP $\hat{\vz} = \arg\max_{\vz} \log p^*(\vz)$:
\begin{equation}
\log p^*(\vz) \approx \log p^*(\hat{\vz}) - \tfrac{1}{2}(\vz - \hat{\vz})^\top \mH (\vz - \hat{\vz}),
\label{eq:taylor}
\end{equation}
where $\mH = -\nabla^2_{\vz} \log p^*\big|_{\hat{\vz}}$ is the negative Hessian (precision matrix) at the mode.
The first-order term vanishes at the MAP by definition.
Integrating the resulting Gaussian gives the Laplace approximation to the evidence:
\begin{equation}
\log \mathcal{Z} \approx \log \mathcal{L}(\data \mid \hat{\vz}) + \log \pi(\hat{\vz}) + \tfrac{n}{2}\log(2\pi) - \tfrac{1}{2}\log\det\mH,
\label{eq:log_laplace}
\end{equation}
where $n = \dim(\vz)$.
This is exact when $p^*$ is Gaussian; the approximation quality degrades for heavy-tailed, skewed, or multimodal posteriors.

\subsection{Nested Sampling}
\label{subsec:ns}

Nested sampling~\citep{Skilling2006} computes $\mathcal{Z}$ by transforming \cref{eq:evidence} into a one-dimensional integral over the prior volume $X(\ell) = \int \pi(\vtheta)\,\mathbf{1}[\mathcal{L}(\vtheta) > \ell]\,d\vtheta$:
\begin{equation}
\mathcal{Z} = \int_0^1 \mathcal{L}(X)\,dX,
\end{equation}
where $\mathcal{L}(X)$ is the likelihood at prior volume $X$.
The algorithm maintains $m$ live particles distributed according to the prior constrained to the current likelihood threshold $\ell^*$.
At each iteration it removes the particle with the lowest likelihood, records its contribution to $\mathcal{Z}$, and replaces it with a new draw from the constrained prior $\pi(\vtheta)\,\mathbf{1}[\mathcal{L}(\vtheta) > \ell^*]$.

The total computational cost at fixed precision in $\log\mathcal{Z}$ scales as
\begin{equation}
T_\mathrm{NS} \propto T_{\mathcal{L}} \times m \times f_\mathrm{sampler} \times D_{\mathrm{KL}},
\label{eq:ns_cost}
\end{equation}
where $T_\mathcal{L}$ is the cost per likelihood evaluation, $m \sim \mathcal{O}(D)$ is the required number of live points, $f_\mathrm{sampler} \sim \mathcal{O}(D)$ is the number of inner sampler steps required for decorrelation, and $D_\mathrm{KL}$ is the Kullback--Leibler divergence from prior to posterior, which scales as $\mathcal{O}(D)$ for typical models \citep{petrosyan2023supernest}.
Together these give a cubic scaling $\mathcal{O}(D^3)$ in parameter dimension $D$~\citep{Yallup2026}.
This makes full nested sampling infeasible for $D \gtrsim 100$ in practice, motivating ALCS as a way to explore only the collapsed space, reducing the precision ($m$), decorrelation scale ($f$) and prior contraction ($D_\mathrm{KL}$) required for robust evidence calculation.

\section{Method}
\label{sec:method}

\subsection{Problem Setup}
\label{subsec:setup}

We consider models in which the parameters split naturally into two groups: a set $\vtheta \in \mathbb{R}^{d_\theta}$ of \emph{parameters of interest} and a set $\vz \in \mathbb{R}^{d_z}$ of \emph{latent parameters} to be marginalised.
Common examples include per-object nuisance parameters in hierarchical analyses~\citep{GelmanBDA2013}, discretised field values in field-level inference~\citep{Jasche2013}, and latent function values in Gaussian process models~\citep{rasmussen_williams_2006}. Whilst there are some notable applications where full joint sampling of parameter space is necessary, for example Bayesian Neural Network posteriors~\citep{pmlr-v119-wenzel20a}, there are many applications where the factorisation either holds explicitly or approximately.
We assume the prior factorises as $\pi(\vtheta, \vz) = \pi(\vtheta)\,\pi(\vz \mid \vtheta)$, and we wish to compute
\begin{equation}
\mathcal{Z} = \int \left[ \int \mathcal{L}(\data \mid \vtheta, \vz)\,\pi(\vz \mid \vtheta)\,d\vz \right] \pi(\vtheta)\,d\vtheta
= \int \mathcal{L}_\mathrm{marg}(\vtheta)\,\pi(\vtheta)\,d\vtheta,
\label{eq:evidence_split}
\end{equation}
where we define the \emph{latent-marginalised likelihood}
\begin{equation}
\mathcal{L}_\mathrm{marg}(\vtheta) = \int \mathcal{L}(\data \mid \vtheta, \vz)\,\pi(\vz \mid \vtheta)\,d\vz.
\label{eq:lmarg}
\end{equation}
If $\mathcal{L}_\mathrm{marg}(\vtheta)$ can be evaluated, an outer sampler can operate in the low-dimensional $\vtheta$ space, simplifying the geometry and dimensionality the sampler needs to explore and thus reducing the number of likelihood evaluations needed to draw mixed, de-correlated samples. Alternative approaches learn marginal posteriors from existing posterior samples via normalising flows~\citep{Bevins2022a,Bevins2022b}; ALCS instead evaluates $\mathcal{L}_\mathrm{marg}(\vtheta)$ directly at each $\vtheta$.

ALCS approximates $\mathcal{L}_\mathrm{marg}(\vtheta)$ using local gradient information about $\vz$ at each fixed $\vtheta$. To implement the \emph{outer} nested sampling loop we use the nested slice sampling implementation of \citet{NSSyallup}, which uses short slice sampling~\citep{Neal2003} to propagate Markov chains in $\vtheta$ space, targeting the marginalised evidence integral of \cref{eq:evidence_split}. This framework evaluates a batch of $k$ deleted points in parallel, synchronising at the likelihood level, so that all ALCS calls within a batch are vectorised across modern hardware accelerators.

\subsection{The ALCS Algorithm}
\label{subsec:ALCS}

At a fixed $\vtheta$, define the joint unnormalised conditional posterior of the latents,
\begin{equation}
p^*(\vz \mid \data, \vtheta) = \mathcal{L}(\data \mid \vtheta, \vz)\,\pi(\vz \mid \vtheta).
\end{equation}
ALCS proceeds in two steps for each nested sampling likelihood evaluation:

\paragraph{Step 1: Optimisation.}
Find the conditional MAP of the latents,
\begin{equation}
\hat{\vz}(\vtheta) = \underset{\vz}{\arg\max}\left[\log \mathcal{L}(\data \mid \vtheta, \vz) + \log \pi(\vz \mid \vtheta)\right],
\label{eq:map}
\end{equation}
using a gradient-based optimiser (e.g.\ L-BFGS), with gradients provided by automatic differentiation.

\paragraph{Step 2: Laplace approximation.}
Compute the negative Hessian at the MAP,
\begin{equation}
\mH(\vtheta) = -\nabla^2_{\vz}\left[\log \mathcal{L}(\data \mid \vtheta, \vz) + \log \pi(\vz \mid \vtheta)\right]\Big|_{\vz = \hat{\vz}(\vtheta)},
\label{eq:hessian}
\end{equation}
which is the precision matrix when $p^*$ is Gaussian, and apply \cref{eq:log_laplace} to obtain marginalised log-likelihood:
\begin{equation}
\log \mathcal{L}_\mathrm{ALCS}(\vtheta) = \log \mathcal{L}(\data \mid \vtheta, \hat{\vz}) + \log \pi(\hat{\vz} \mid \vtheta) + \tfrac{d_z}{2}\log(2\pi) - \tfrac{1}{2}\log\det\mH(\vtheta).
\label{eq:ALCS}
\end{equation}
This expression is then used as the likelihood for the nested sampling outer loop over $\vtheta$.

The Laplace approximation is exact when $p^*(\vz \mid \vtheta)$ is Gaussian; it is a good approximation when the latent posterior is approximately unimodal and well-approximated by a Gaussian near its peak.
In \cref{sec:experiments} we demonstrate both regimes; where the approximation is highly accurate and where it breaks down.

\begin{algorithm}[t]
\caption{ALCS: marginalised log-likelihood evaluation at $\vtheta$}
\label{alg:ALCS}
\begin{algorithmic}[1]
\Require Differentiable log-joint $\log p(\data, \vz \mid \vtheta)$; current hyperparameters $\vtheta$; warm-start $(\hat{\vz}_\mathrm{prev}, H_\mathrm{prev})$ (optional)
\Ensure Marginalised log-likelihood $\log \mathcal{L}_\mathrm{ALCS}(\vtheta)$
\State \textbf{Optimise} latents: $\hat{\vz}(\vtheta) \leftarrow \arg\max_{\vz}\,\log p(\data, \vz \mid \vtheta)$
  \quad \textit{(L-BFGS; initialise from $\hat{\vz}_\mathrm{prev}$ if available)}
\State \textbf{Compute Hessian}: $\mH(\vtheta) \leftarrow -\nabla^2_{\vz}\log p(\data, \vz \mid \vtheta)\big|_{\hat{\vz}}$
  \quad \textit{(full via \texttt{jax.hessian}, or sparse via HVPs; parallelised with \texttt{jax.vmap})}
\State \textbf{Log-determinant}: $\ell_H \leftarrow \tfrac{1}{2}\log\det\mH(\vtheta)$
  \quad \textit{(Cholesky; $\mathcal{O}(d_z^3)$ dense or $\mathcal{O}(d_z)$ tridiagonal)}
\State \textbf{Return}:
\[
  \log\mathcal{L}_\mathrm{ALCS}(\vtheta)
  = \log p(\data, \hat{\vz} \mid \vtheta)
  + \tfrac{d_z}{2}\log(2\pi)
  - \ell_H
\]
\State \textbf{Cache} $(\hat{\vz}, \mH)$ as warm start for the next call \Comment{optional}
\end{algorithmic}
\end{algorithm}

\noindent
\Cref{alg:ALCS} describes the procedure run at every evaluation of $\mathcal{L}_\mathrm{marg}(\vtheta)$ by the outer sampler; for nested sampling this is once per dead point, and calls are independent across live points and therefore parallelisable using \texttt{jax.vmap} for JAX-native optimisers such as \texttt{jaxopt}~\citep{Blondel2022}.
Posterior samples in the full joint space $(\vtheta, \vz)$ can be recovered post-hoc from the NS output; see \Cref{app:posterior_recovery}.

\subsection{Computational Scaling}
\label{subsec:scaling}

The cost of one ALCS likelihood evaluation at $\vtheta$ consists of:
\begin{itemize}
  \item Optimisation: $N_\mathrm{opt}$ gradient evaluations, each costing $\mathcal{O}(T_\mathcal{L} + T_\pi)$, where $T_\mathcal{L}$ is the cost of evaluating the likelihood and $T_\pi$ the prior. In practice $N_\mathrm{opt}$ is small and nearly dimension-independent for well-conditioned problems~\citep{LiuNocedal1989}.
  \item Hessian: $\mathcal{O}(d_z)$ gradient evaluations via \texttt{jax.hessian} (forward-over-reverse), giving cost $\mathcal{O}(d_z \times T_\mathcal{L})$.
  \item Log-determinant: $\mathcal{O}(d_z^3)$ via Cholesky factorisation, or $\mathcal{O}(d_z)$ when the Hessian is block-diagonal or banded (see \cref{app:hessian}).
\end{itemize}

The total nested sampling cost is then
\begin{equation}
T_\mathrm{ALCS} \propto T_{\mathcal{L}_\mathrm{ALCS}} \times m_\theta \times f_\theta \times D_{\mathrm{KL},\theta},
\label{eq:ALCS_cost}
\end{equation}
where $m_\theta$, $f_\theta$, and $D_{\mathrm{KL},\theta}$ all now scale only with the hyperparameter dimension $d_\theta$.
This scales as $\mathcal{O}(d_\theta^3 \times d_z)$, which for $d_\theta \ll d_z$ represents a substantial improvement over the $\mathcal{O}((d_\theta + d_z)^3)$ cost of joint nested sampling.
Implementation details (JAX setup, prior whitening, warm-start preconditioning, block-diagonal factorisation, and structured Hessian computation) are described in \Cref{app:implementation}.

\paragraph{Assumptions and Scope.}
The Laplace approximation requires that $p(\vz \mid \vtheta, \data)$ be approximately unimodal and well-approximated by a Gaussian near its mode, for all $\vtheta$ in the posterior support.
If the latent conditional is multimodal, the MAP optimisation may converge to a local rather than global mode, yielding an incorrect Hessian and a biased evidence estimate.
Due to the local nature of the ALCS estimate this restriction is non-negotiable, but it is not peculiar to ALCS: multimodal distributions in high latent dimension are intractable for essentially all inference methods.

Any method that scales to $d_z \sim 10^4$ must make strong structural assumptions about the latent space. ALCS makes its assumption explicit: conditioned on $\vtheta$, it propagates $\vz$ uncertainty into uncertainty on $\vtheta$ and $\mathcal{Z}$ at quadratic order.

From the above computational scaling we infer the scope of this method by considering the bottlenecks of each required step. For the optimisation and $N_\text{opt}$, ALCS is designed for near-Gaussian latent spaces, and as such we are primarily targeting quadratic surfaces. Although optimisation in such spaces scales well~\citep{LiuNocedal1989,Zhang2022pathfinder}, it breaks down in anisotropic spaces, which can be monitored via the condition number of the Hessian \citep{LiuNocedal1989}. The Hessian computation provides the largest cost for most of the likelihoods considered here (where typically $d_z \gg N_\mathrm{opt}$), and in ultra-high dimension may become numerically unstable. Finally the computation of the volume contribution (the Hessian's log-determinant) provides a barrier to field level ($d_z \sim \mathcal{O}(10^5)$ analyses, and will require determinant estimators or even greater structural assumptions about the Hessian.

We also note that the outer slice sampling moves in $\vtheta$ have some variance in trajectory length per sample, however as was established in~\citet{NSSyallup} this does not seriously impact the parallelism.

\subsection{Student-$t$ Extension}
\label{subsec:student_t}

When the latent posterior $p(\vz \mid \vtheta, \data)$ is heavier-tailed than Gaussian, for example, when the prior on $\vz$ is itself a Student-$t$ distribution, the Gaussian Laplace approximation will systematically underestimate the evidence.
This can be corrected by replacing the Gaussian with a richer local approximation.

ALCS relies on comparing the unnormalised posterior mode to the known mode of a best-fit Gaussian distribution, matched to the local shape at that point. This generalises: any normalised distribution whose shape parameters can be inferred from local derivatives of $\log p^*$ at the MAP (a skew-normal from the third derivative, a $\chi^2$ family from the asymmetry, and so on) yields a valid replacement for the Gaussian normalising constant in \cref{eq:ALCS}, with no other change to the framework.

In this work we consider using 1D Student-$t$ distribution in the Cholesky-whitened basis $\bm{w} = L(\vz - \hat{\vz})$ (where $\mH = LL^\top$), treating each whitened direction $j$ independently.
The degrees of freedom $\nu_j$ are estimated from the fourth derivative at the MAP:
\begin{equation}
  \hat\kappa_j = \frac{3\,f''''_j(\hat{\vz})}{[f''_j(\hat{\vz})]^2} - 3, \qquad
  \hat\nu_j = 4 + \frac{6}{\hat\kappa_j},
  \label{eq:kurtosis-4d}
\end{equation}
where $f_j$ denotes the log-posterior restricted to whitened direction $j$ and $\hat\kappa_j$ is the estimated excess kurtosis.
Substituting the Student-$t(\nu_j)$ normalising constant per whitened direction in place of the Gaussian gives
\begin{equation}
\log \mathcal{L}_\mathrm{St}(\vtheta)
= \log \mathcal{L}_\mathrm{ALCS}(\vtheta)
- \tfrac{d_z}{2}\log(2\pi)
+ \sum_{j=1}^{d_z} \log q_j(\nu_j),
\label{eq:student_full}
\end{equation}
where $\log q_j(\nu_j) = \log\Gamma\!\bigl(\tfrac{\nu_j+1}{2}\bigr) - \log\Gamma\!\bigl(\tfrac{\nu_j}{2}\bigr) - \tfrac{1}{2}\log[\pi(\nu_j{+}1)]$ is the log normalising constant of a unit Student-$t(\nu_j)$.
As $\nu_j \to \infty$, $\log q_j \to \tfrac{1}{2}\log(2\pi)$ and \cref{eq:student_full} reduces to \cref{eq:ALCS}.
The derivation is given in \cref{app:derivation}.
Estimating $\hat\nu_j$ requires only two additional scalar autodiff evaluations per whitened direction at the already-computed MAP so is entirely local with no additional sampling.
We show in \cref{sec:experiments} that the fourth-derivative estimator reduces the evidence error when the latent posterior is genuinely heavy-tailed.

This remains a \emph{local} correction: it cannot fix approximation failures that arise from non-local structure, and any such method would require sampling or exploring the latent space, which this method explicitly avoids. When the posterior is Gaussian the correction correctly vanishes, albeit with the extra cost of the fourth-derivative evaluations.
One might expect the evidence itself to select between the Gaussian and Student-$t$ approximations, as Bayesian model comparison normally would. However, this is not safe here: because both estimates are anchored at the same MAP and infer tail behaviour only from local derivatives, the Student-$t$ correction can both under- and over-estimate the evidence depending on whether the true tails are heavier or lighter than the fitted $\hat\nu_j$ implies. The IS ESS diagnostic (\cref{app:is_diagnostic}) provides a more reliable indicator of whether the richer approximation is warranted.

\section{Experiments}
\label{sec:experiments}

We evaluate ALCS on a suite of benchmarks spanning a range of latent posterior structures.
Three primary experiments establish the core properties: a hierarchical supernova model (demonstrating exact-Gaussian accuracy and $D=25{,}600$ scaling), a Student-$t$ extension for heavy-tailed latents (demonstrating that ALCS's autodiff framework extends beyond simple Gaussian approximations), and a tanh funnel model (mapping the failure mode when latent geometry is highly non-Gaussian).
A further six models from the \texttt{inference\_gym} benchmark suite~\citep{Sountsov2020} provide systematic coverage of hierarchical, time-series, and discrete-likelihood structures.
All experiments use the BlackJAX \citep{Cabezas2024} nested slice sampling implementation as the basis for the outer loop~\citep{Yallup2026} with hyperparameter settings of $m = 500$ live points, $k = 100$ batch-deleted points per iteration, $5$ repeated slice sampling steps and terminating the algorithm when the standard condition of $\log \mathcal{Z}_\text{live} - \log \hat{\mathcal{Z}} < -3$ is met. MAP estimates $\hat{\vtheta}$ are obtained by L-BFGS; Hessians are computed via \texttt{jax.hessian} on an NVIDIA H200 GPU (Isambard-AI). Each experiment is run $R = 5$ times with different random seeds on the same data; error bars show mean $\pm$ standard deviation across runs. The full NS ground truth uses the same BlackJAX implementation run jointly over all parameters $(\vtheta, \vz)$; $\hat{\sigma}$ denotes its internal uncertainty estimate.

\subsection{Supernova Cosmology}
\label{subsec:sne}

\paragraph{Model.}
We use a hierarchical supernova type Ia model based on the Tripp formula~\citep{tripp1998}: each of $N$ supernovae has latent stretch and colour parameters $\vz_i = (x_{1,i}, c_i)$ whose contribution to the distance modulus depends on the cosmology $\vtheta$ under test, so ALCS propagates the per-object measurement uncertainty at each $\vtheta$ automatically.
The Gaussian likelihood and Gaussian conditional prior $\pi(\vz_i \mid \vtheta)$ admit an analytic marginalisation, providing an exact $\mathcal{L}_\text{marg}(\vtheta)$ that serves as the reference outer loop likelihood against which ALCS is compared. While the latents are Gaussian, the outer cosmological posterior $p(\vtheta \mid \data)$ has a complex degeneracy, owing to the non-linear forward model from $\vtheta$ to the observables. Full model specification is given in \Cref{app:exp_sne}.

\paragraph{Test 1: scaling with $N$.}
We fix $d_z = 2$ and scale $N \in \{64, 128, 256, 512, 1024, 2048\}$, giving total latent dimension $D = 2N$ up to 4096. The ALCS Hessian is block-diagonal with $N$ independent $2 \times 2$ blocks, computed with \texttt{jax.vmap} over all objects simultaneously.

\paragraph{Test 2: scaling with latent dimension.}
We fix $N = 100$ and scale the per-object latent dimension $d_{z,\text{block}} \in \{2, 4, 8, 16, 32, 64, 128, 256\}$, giving total $D = N \cdot d_{z,\text{block}}$ up to $25{,}600$.

\begin{figure}[t]
\centering
\includegraphics[width=\textwidth]{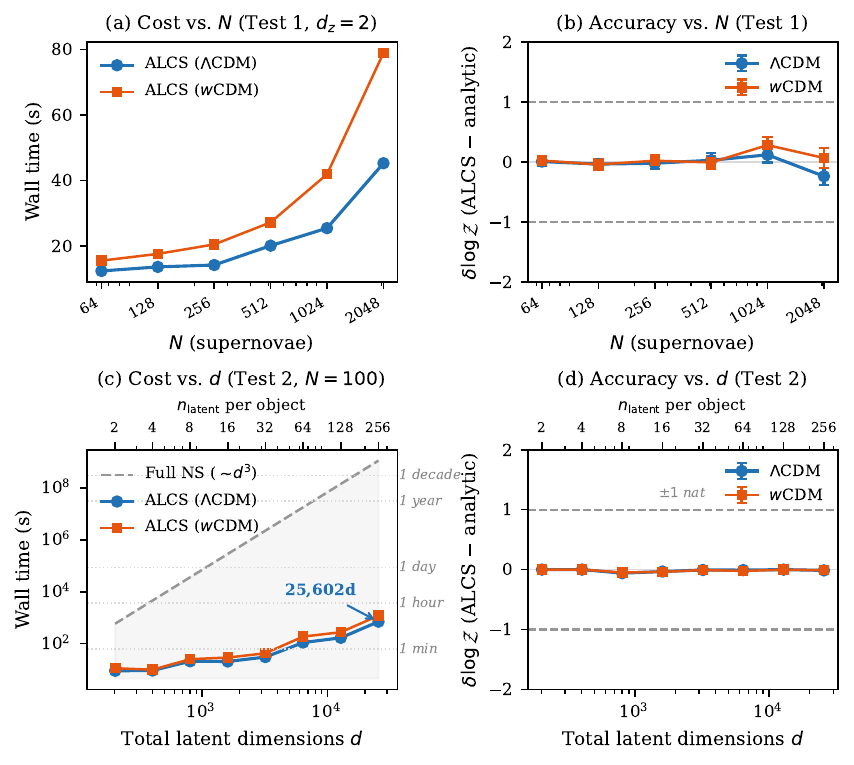}
\caption{Supernova cosmology scaling for $\Lambda$CDM and $w$CDM.
\textbf{(a--b)} Test 1: wall time and evidence error $\delta\log\mathcal{Z}$ vs $N$ at fixed $d_{z,\text{block}}=2$.
\textbf{(c--d)} Test 2: wall time (log--log) and $\delta\log\mathcal{Z}$ vs total latent dimension $D$ at fixed $N=100$; the $D^3$ full-NS reference is projected.
Top axis in (c) shows latents per object.}
\label{fig:sne_scaling}
\end{figure}

Results are shown in \Cref{fig:sne_scaling}.

\paragraph{Results.}
\textit{Test 1.}
Wall time grows from $12\,\mathrm{s}$ ($N=64$) to $45\,\mathrm{s}$ ($N=2048$) for $\Lambda$CDM (\Cref{fig:sne_scaling}a).
As shown in \Cref{app:dkl}, this growth is entirely attributable to the increasing $D_\mathrm{KL}(\vtheta)$ as more supernovae constrain the cosmological parameters; the per-call cost $t_\mathrm{single} \approx 1.2\,\mathrm{ms}$ is flat across all $N$ (\Cref{fig:dkl}).
Evidence errors remain below $0.25$ nats for all $N$ and both models, with most points within $\hat{\sigma}$ (\Cref{fig:sne_scaling}b).

\textit{Test 2.}
ALCS wall time grows from $9\,\mathrm{s}$ at $D=200$ to $693\,\mathrm{s}$ (${\approx}12\,\mathrm{min}$) at $D=25{,}602$, while the projected full-NS $D^3$ reference reaches ${\approx}37$ years at the same scale (\Cref{fig:sne_scaling}c).
The evidence error $|\delta\log\mathcal{Z}|$ never exceeds $0.06$ nats across all eight scales and both models, confirming that the Gaussian Laplace approximation is finding the exact solution for this model (\Cref{fig:sne_scaling}d).


\subsection{Student-$t$ Extension: Beyond the Gaussian Assumption}
\label{subsec:student_t_exp}

The ALCS framework makes it straightforward to match higher-order moments: we estimate the degrees of freedom $\nu_j$ from the fourth derivative of the log-posterior at the MAP (\Cref{subsec:student_t}) and use the Student-$t$ marginalised likelihood formula (\Cref{eq:student_full}).

\paragraph{Model.}
We use a hierarchical model with $N_\text{obj}$ scalar latents:
\begin{equation*}
  \vtheta = (\mu, \log\sigma) \sim \mathrm{Uniform},\quad
  z_i \mid \vtheta \sim \mathrm{Student}\text{-}t(\nu_\text{true}{=}5,\,\mu,\,\sigma),\quad
  y_i \mid z_i \sim \mathcal{N}(z_i,\,\sigma_\text{obs}^2{=}1).
\end{equation*}
The heavy-tailed prior creates non-Gaussian latent conditionals $p(z_i \mid \vtheta, y_i)$, but the Gaussian likelihood softens the tails relative to the $\nu{=}5$ prior.
We compare Gaussian ALCS, Student-$t$ ALCS (per-observation $\hat\nu$ from the fourth derivative), and full NS ground truth across $N_\text{obj} \in \{10, 20, 50, 100, 150\}$.
This is a deliberately synthetic setup: the Student-$t$ prior is chosen precisely to guarantee lightly non-Gaussian latent conditionals, providing a controlled test of whether the fourth-derivative correction captures the right effective $\nu$, not a claim that such priors are typical in practice.
Prior bounds and data generation details are in \Cref{app:exp_student}.

\begin{figure}[t]
\centering
\includegraphics[width=\textwidth]{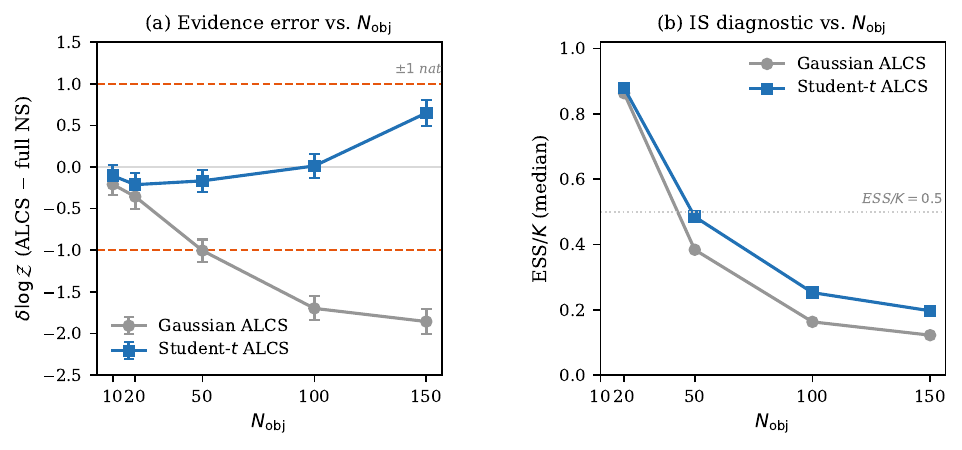}
\caption{Student-$t$ extension ($\nu_\text{true}=5$, $N_\text{obj}\in\{10,20,50,100,150\}$).
(a) Evidence error $\delta\log\mathcal{Z}$ for Gaussian (grey) and Student-$t$ (blue) ALCS.
(b) IS ESS$/K$ (median over $M{=}200$ posterior samples). See \Cref{tab:student} for numerical values.}
\label{fig:student}
\end{figure}

\paragraph{Results.}
Gaussian ALCS accumulates an evidence error of $-1.00$ nats at $N_\text{obj} = 50$ and $-1.86$ nats at $N_\text{obj} = 150$ (\Cref{fig:student}a).
The Student-$t$ correction substantially reduces this: at $N_\text{obj} = 50$ the error falls to $-0.17$ nats ($83\%$ improvement), and at $N_\text{obj} = 100$ the corrected estimate agrees with full NS to within $0.01$ nats.
\Cref{fig:student}b shows the IS ESS$/K$ diagnostic (\Cref{app:is_diagnostic}) for both proposals across $N_\text{obj}$.
The Student-$t$ proposal consistently achieves higher ESS$/K$ than the Gaussian (e.g.\ $0.49$ vs $0.38$ at $N_\text{obj} = 50$), confirming that the fourth-derivative $\hat\nu$ estimator captures the correct effective tail weight and that the Student-$t$ extension provides a better marginalised likelihood for heavy-tailed latent posteriors.
Full evidence estimates and ESS$/K$ values are tabulated in \Cref{tab:student}.

\subsection{Neal's Funnel: Scope of the Gaussian Assumption}
\label{subsec:funnel}

\begin{figure}[t]
\centering
\includegraphics[width=0.75\textwidth]{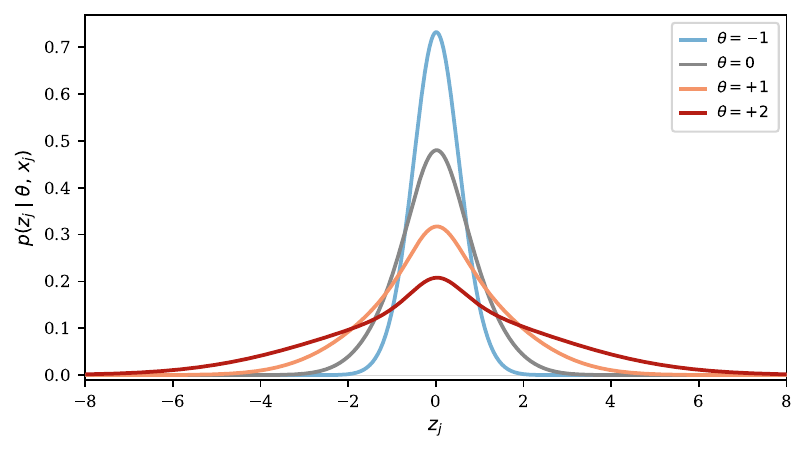}
\caption{Latent conditional $p(z_j\mid\theta,x_j)$ for $\theta\in\{-1,0,+1,+2\}$.
At $\theta<0$ the posterior is near-Gaussian; as $\theta$ increases $\tanh(z_j)$ saturates and flat shoulders develop.}
\label{fig:funnel_latent}
\end{figure}

Neal's funnel~\citep{Neal2003} is a classically difficult sampling problem for gradient-based samplers in particular, necessitating the use of reparameterisations to neutralise this geometric pathology~\citep{gorinova2019automaticreparameterisationprobabilisticprograms}. We consider a further challenge on top of this, incorporating non-Gaussianity in the latent variables, a challenging test case from field-level inference in cosmology~\citep{Millea:2021had}.
As it is sufficient to display the failure, we use a version with only $d_\theta = 1$ hyperparameter, $J = 10$ latent variables, and a tanh-compressed observation model:
\begin{equation}
  \theta \sim \mathcal{N}(0, 9),\quad
  z_j \mid \theta \sim \mathcal{N}(0, e^\theta),\quad
  x_j \mid z_j \sim \mathcal{N}(\tanh(z_j),\, 1).
  \label{eq:tanh_funnel}
\end{equation}
Data are generated once at $\theta_\text{true} = 0$.
The tanh observation model creates a directional failure: for $\theta > 0$ the prior on $z_j$ is wide, so large $|z_j|$ values are accessible, but $\tanh(z_j)$ saturates at $\pm 1$.
The true likelihood $\mathcal{L}_\text{true}(\theta)$ is therefore approximately flat for $\theta \gg 0$ (the data can always be fit by pushing $z_j$ into the flat region of tanh), while the Gaussian Laplace approximation has a quadratic well centred on the MAP and falls off steeply.
At $\theta < 0$ the prior forces $z_j \approx 0$ and the posterior is approximately Gaussian, so ALCS is accurate there.
\Cref{fig:funnel_latent} shows the latent conditional $p(z_j\mid\theta,x_j)$ for representative values of $\theta$: as $\theta$ increases the distribution widens and develops flat shoulders that no Gaussian centred on the MAP can capture.

\begin{figure}[t]
\centering
\includegraphics[width=\textwidth]{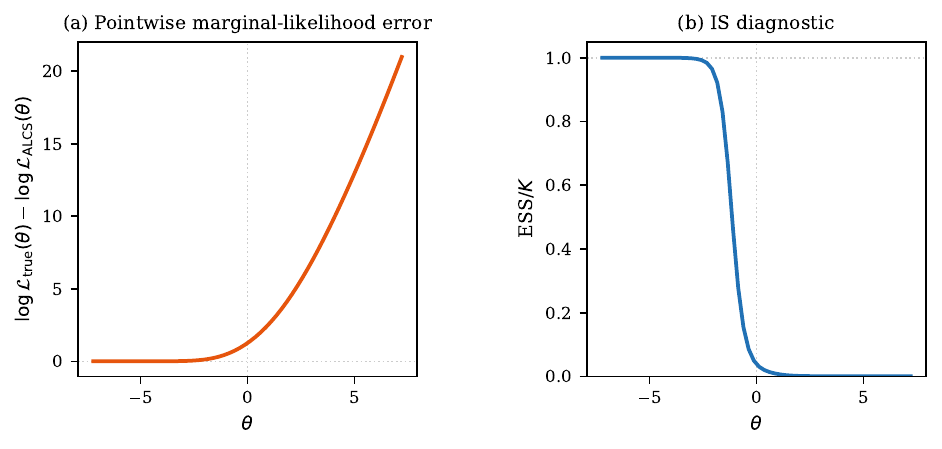}
\caption{Tanh funnel ($J=10$).
\textbf{(a)} Pointwise evidence error vs $\theta$: near zero for $\theta<0$, growing to ${\sim}20$ nats by $\theta=3$.
\textbf{(b)} IS ESS$/K$ at each $\theta$ ($K{=}5000$): ${\approx}1$ for $\theta<0$, dropping to ${\ll}0.01$ for $\theta>0$, localising the approximation failure.}
\label{fig:funnel}
\end{figure}

\paragraph{Results.}
Full NS: $\log\hat{\mathcal{Z}} = -15.94 \pm 0.05$
Gaussian ALCS: $\log\hat{\mathcal{Z}} = -16.68 \pm 0.04$ (error $-0.74$ nats).
The error is ${\sim}18\hat{\sigma}$, confirming ALCS is not appropriate here.
Panel~(a) of \Cref{fig:funnel} illustrates the mechanism: at $\theta < 0$ the error is near zero, whereas for $\theta > 0$ the approximation degrades rapidly as tanh saturates.

The Student-$t$ correction of \Cref{subsec:student_t} cannot fix this failure: the flat shoulders arise from the far tails of the latent posterior, which are non-local and entirely invisible to any gradient evaluation at the MAP.
Near the MAP at $\theta \gg 0$, the posterior looks Gaussian, the kurtosis estimator returns $\hat\nu \approx \infty$ because the failure is in a region of $z_j$-space far from the mode.
This is a qualitatively different failure from heavy tails, and no local higher-derivative correction can address it.

The IS ESS$/K$ diagnostic~(\Cref{app:is_diagnostic}) shows a correspondence between the non-Gaussianity and the ALCS error; evaluated across $\theta$-space, it shows ESS$/K \approx 1$ at $\theta < 0$ (Gaussian regime, approximation valid) and ESS$/K \ll 1$ at $\theta \gg 0$ (tanh saturation, approximation invalid).
This spatial view identifies exactly the region where ALCS breaks down and where a more flexible method would be needed.

We note two important limiting cases.
First, replacing the tanh observation model with a linear-Gaussian one, $x_j \mid z_j \sim \mathcal{N}(z_j, 1)$, yields an exactly Gaussian latent conditional and ALCS is exact for all $J$ and all $\theta$: the standard Neal's funnel~\citep{Neal2003} is not a challenging case.
Second, because the $J$ latents are conditionally independent given $\theta$, the per-latent evidence error is independent of $J$, and the total evidence error in the tanh case accumulates as $\mathcal{O}(J)$.
This is an important practical consideration: the tanh failure mode worsens linearly with the number of latents, reinforcing that the Gaussian assumption must be checked per latent conditional before applying ALCS to large-$J$ models.

\subsection{Inference Gym Benchmarks}
\label{subsec:inference_gym}

We evaluate ALCS on six models from the \emph{inference gym} benchmark suite~\citep{Sountsov2020}, chosen to span a range of latent structure: Gaussian-Latent models (Eight Schools, Radon, Brownian Motion), a rich-data non-conjugate model (LGCP), a discrete-likelihood model (IRT), and a non-linear time series (Stochastic Volatility).
As a wall-time baseline we run No-U-Turn Sampling (NUTS)~\citep{Hoffman2014} via BlackJAX on the same H200 GPU, using 4 chains with 2000 warmup and 2000 sampling steps on the full joint parameter space $(\vtheta, \vz)$.
We note that this comparison is not entirely fair: ALCS evaluates all live points concurrently via \texttt{jax.vmap} and is therefore natively parallel on GPU, whereas NUTS chains are inherently sequential in their trajectory and GPU-efficient variants remain an active area of investigation~\citep{pmlr-v130-hoffman21a, rioudurand2023adaptivetuningmetropolisadjusted}.
The wall-time speedups should therefore be read as indicative rather than definitive.
\Cref{tab:inference_gym} summarises the evidence accuracy; results are discussed below. 
Full details of all six models are given in \Cref{app:exp_gym}.

\begin{table*}[t]
\centering
\small
\caption{ALCS on six inference gym benchmarks. $\delta = \log\mathcal{Z}_\text{ALCS} - \log\mathcal{Z}_\text{ref}$ (mean $\pm$ std over 5 runs); ESS$/K$ is the median IS diagnostic (\Cref{app:is_diagnostic}) over $M{=}200$ posterior samples.
$^\dagger$Full joint NS over 501 dimensions is intractable; $\vtheta$ posterior validated against Stan MCMC instead.}
\label{tab:inference_gym}
\smallskip
\begin{tabular}{lrrrrrr}
\toprule
Model & $d_\theta$ & $d_z$ & $\delta$ & $\hat\sigma$ & ESS$/K$ & Ref \\
\midrule
Eight Schools           &  2 &    8 & $+0.08 \pm 0.04$   & $0.05$ & $1.00$ & analytic \\
Radon ($J=85$)          &  4 &   85 & $\phantom{+}0.00 \pm 0.16$ & $0.17$ & $1.00$ & analytic \\
Brownian Motion ($T=50$)&  1 &   50 & $+0.06 \pm 0.07$   & $0.07$ & $1.00$ & Kalman   \\
LGCP ($M=100$)          &  2 &  100 & $+0.12 \pm 0.07$   & $0.08$ & $0.71$ & full NS  \\
SV, SP500 ($T=100$)     &  3 &  100 & $-0.32 \pm 0.18$   & $0.14$ & $0.67$ & full NS  \\
IRT ($N_s=400$)$^\dagger$ &  1 & 500 & ---              & ---    & $0.10$ & Stan$^\dagger$ \\
\bottomrule
\end{tabular}
\end{table*}

\begin{table}[t]
\centering
\caption{NUTS vs ALCS wall time on the six inference gym benchmarks (4 chains, 2000 warmup\,+\,2000 samples).
Min-ESS is the minimum bulk ESS across all parameters (ArviZ). NC = non-centred parameterisation~\citep{Papaspiliopoulos2007}.
$^\dagger$Tasks are not directly comparable: NUTS samples the full 501D joint; ALCS marginalises 500 latents and additionally computes $\log\mathcal{Z}$.}
\label{tab:nuts_comparison}
\smallskip
\begin{tabular}{lrrrrrr}
\toprule
Model & $d_z$ & Param. & Min-ESS & $t_\text{NUTS}$ (s) & $t_\text{ALCS}$ (s) & Speedup \\
\midrule
Eight Schools      &    8 & NC      & $1149$ &    44 &     3 & $15\times$ \\
Radon ($J=85$)     &   85 & NC      & $3367$ &    62 &     4 & $17\times$ \\
Brownian ($T=50$)  &   50 & centred & $2280$ &    52 &     4 & $13\times$ \\
LGCP ($M=100$)     &  100 & centred & $7994$ &    91 &    33 & $\phantom{0}3\times$  \\
SV ($T=100$)       &  100 & NC      &  $156$ &  1160 &   329 & $\phantom{0}4\times$  \\
IRT ($N_s=400$)    &  500 & centred &  $314$ &   102 &  3480 & $0.03\times^\dagger$ \\
\bottomrule
\end{tabular}
\\[4pt]
\small$^\dagger$NUTS here samples the full 501D joint; ALCS runs NS over $d_\theta=1$ only and additionally computes $\log\mathcal{Z}$.
\end{table}

\paragraph{Near-Gaussian models.}
ALCS is exact for Eight Schools, Radon, and Brownian Motion ($\delta = 0.000$, $-0.003$, $-0.027$ nats respectively): when the latent conditional is Gaussian the approximation is exact, and these results confirm that the optimisation and Hessian converge correctly for typical problems.
Centred NUTS fails on Eight Schools and Radon (min-ESS\,=\,16 and 4) due to funnel geometry; a non-centred reparameterisation is required (\Cref{tab:nuts_comparison}), whereas ALCS marginalises the latents analytically and needs no such adjustment.

\paragraph{Log-Gaussian Cox Process.}
With ${\sim}100$ events per grid cell the Poisson likelihood is near-Gaussian (CLT), giving $\delta = +0.15$ nats (${\leq}1\hat\sigma$) and a $64\times$ speedup over full $102$D NS ($33\,\mathrm{s}$ vs $35\,\mathrm{min}$).

\paragraph{Stochastic Volatility.}
The log-volatility path has a tridiagonal Hessian (a structure that is exploited for efficiency, details in \cref{app:hessian}), giving $\delta = -0.24$ nats at $T=100$ and graceful scaling to $T=2516$ ($t_\text{total} = 1883\,\mathrm{s}$; full scaling data in \Cref{tab:sv_scaling}).
ALCS is immune to a fundamental limitation of NUTS on this model: with persistence $\hat\beta \approx 0.96$, the AR(1) path creates strong data-induced correlations that no reparameterisation of the latents can remove~\citep{Betancourt2015}, leaving NUTS at min-ESS\,=\,156 even with non-centred parameterisation.

\paragraph{Item Response Theory.}
ALCS marginalises 500 latents and runs NS over the scalar $\mu_\text{ability}$, recovering a posterior consistent with Stan MCMC to within $0.02\hat\sigma$ ($\mu_\text{ability} = 0.078 \pm 0.110$ vs $0.074 \pm 0.112$). Using the converged NUTS samples we considered the Kullback--Leibler divergence~\citep{CoverThomas2006} in each dimension, showing $\sum_i D_\mathrm{KL} = 2.4$ nats over the 500 dimensions (${\sim}0.005$ nats each). This reflects a small near-Gaussian discrepancy which compounds over dimension to bias the evidence; however the recovered $\vtheta$ posterior shows perfect agreement (\cref{fig:irt_theta}).

\section{Conclusion}
\label{sec:conclusion}

We have presented ALCS, a framework for marginalising latent parameters in Bayesian models using automatic differentiation.
By collapsing $\vz$ to a scalar evidence contribution at each evaluation of the marginal likelihood, via MAP optimisation and a Hessian computed by autodiff, ALCS reduces the effective dimensionality seen by an outer sampler from $d_\theta + d_z$ to $d_\theta$, making nested sampling tractable for models with large latent spaces.
The only requirement is a differentiable joint log-posterior; no model-specific gradient or Hessian derivation is needed.

On the supernova cosmology benchmark and Gaussian inference gym problems, where analytic ground truth is available, ALCS achieves sub-$\hat{\sigma}$ evidence accuracy up to $d = 25{,}600$ total latents, with wall time nearly flat in $d$ due to GPU-parallelised vmapped Hessian computation.
The tanh funnel establishes the failure mode: when the latent non-Gaussianity varies strongly with $\vtheta$, the Gaussian Laplace approximation fails in both posterior reconstruction and evidence computation.

The method is most powerful precisely where high-dimensional inference is most needed, in hierarchical models; ALCS is natively amenable to parallel hardware so experiences the greatest speed-up over joint-space samplers for factorising likelihoods.

The IS/ESS diagnostic of \cref{app:is_diagnostic} provides a practical tool for identifying the latter case post-hoc.
Extensions to heavier-tailed approximations (\cref{subsec:student_t}) and structured sparse Hessians (\cref{app:hessian}) widen the applicable model class further, and future work will implement these methods and their extensions on full-scale modern cosmological likelihoods.

\section*{Acknowledgements}

This work was supported by the research environment and infrastructure of the Handley Lab at the University of Cambridge.
TL is supported by the Harding Distinguished Postgraduate Scholars Programme (HDPSP).
The authors acknowledge the use of resources provided by the Isambard-AI National AI Research Resource (AIRR). Isambard-AI is operated by the University of Bristol and is funded by the UK Government's Department for Science, Innovation and Technology (DSIT) via UK Research and Innovation; and the Science and Technology Facilities Council~\citep[ST/AIRR/I-A-I/1023]{McIntoshSmith2024}.
Parts of this work used large language models: Claude (Anthropic) assisted with code and plotting scripts, and Claude and Gemini (Google) assisted with proofreading and manuscript editing.

\bibliography{references}

@techreport{minka2001,
  author      = {Minka, Thomas P.},
  title       = {A Family of Algorithms for Approximate {Bayesian} Inference},
  institution = {Massachusetts Institute of Technology},
  year        = {2001},
  type        = {PhD thesis},
}

@misc{stan2023,
  author       = {{Stan Development Team}},
  title        = {Stan Modeling Language Users Guide and Reference Manual, Version 2.33},
  year         = {2023},
  url          = {https://mc-stan.org},
}

@article{Skilling2006,
  author  = {Skilling, John},
  title   = {Nested sampling for general {B}ayesian computation},
  journal = {Bayesian Anal.},
  volume  = {1},
  number  = {4},
  pages   = {833--859},
  year    = {2006},
  doi     = {10.1214/06-BA127},
}

@article{Ashton2022,
  author  = {Ashton, Gregory and others},
  title   = {Nested sampling for physical scientists},
  journal = {Nat. Rev. Methods Primers},
  volume  = {2},
  pages   = {39},
  year    = {2022},
}

@article{Buchner2023,
  author  = {Buchner, Johannes},
  title   = {Nested sampling methods},
  journal = {Stat. Surv.},
  volume  = {17},
  pages   = {169--215},
  year    = {2023},
}

@article{Yallup2026,
  author        = {Yallup, David and Kroupa, Namu and Handley, Will},
  title         = {Nested Slice Sampling: Vectorized Nested Sampling for {GPU}-Accelerated Inference},
  journal       = {arXiv preprint},
  year          = {2026},
  eprint        = {2601.23252},
  archivePrefix = {arXiv},
}

@article{NS-SwiG,
  author        = {Yallup, David},
  title         = {Nested Sampling with Slice-within-{G}ibbs: Efficient Evidence Calculation for Hierarchical {B}ayesian Models},
  journal       = {arXiv preprint},
  year          = {2026},
  eprint        = {2602.17414},
  archivePrefix = {arXiv},
}

@article{cai2022,
  author        = {Cai, Xiaohao and McEwen, Jason D. and Pereyra, Marcelo},
  title         = {Proximal nested sampling for high-dimensional {B}ayesian model selection},
  journal       = {Stat. Comput.},
  volume        = {32},
  pages         = {87},
  year          = {2022},
  eprint        = {2106.03646},
  archivePrefix = {arXiv},
}

@inproceedings{NSSyallup,
  title     = {Nested Slice Sampling},
  author    = {Yallup, David and Kroupa, Namu and Handley, Will},
  booktitle = {International Conference on Learning Representations (ICLR)},
  year      = {2025},
  url       = {https://openreview.net/forum?id=ekbkMSuPo4},
  note      = {Submitted to ICLR 2024}
}

@incollection{Neal2011,
  author    = {Neal, Radford M.},
  title     = {{MCMC} using {H}amiltonian dynamics},
  booktitle = {Handbook of {M}arkov Chain {M}onte {C}arlo},
  editor    = {Brooks, Steve and Gelman, Andrew and Jones, Galin and Meng, Xiao-Li},
  publisher = {Chapman and Hall/{CRC}},
  year      = {2011},
}

@article{Hoffman2014,
  author  = {Hoffman, Matthew D. and Gelman, Andrew},
  title   = {The {No-U-Turn} Sampler: Adaptively Setting Path Lengths in {H}amiltonian {M}onte {C}arlo},
  journal = {J. Mach. Learn. Res.},
  volume  = {15},
  pages   = {1593--1623},
  year    = {2014},
}

@article{Neal2003,
  author  = {Neal, Radford M.},
  title   = {Slice sampling},
  journal = {Ann. Stat.},
  volume  = {31},
  number  = {3},
  pages   = {705--767},
  year    = {2003},
}

@article{gelman_simulating_1998,
  author  = {Gelman, Andrew and Meng, Xiao-Li},
  title   = {Simulating normalizing constants: From importance sampling to bridge sampling to path sampling},
  journal = {Stat. Sci.},
  volume  = {13},
  number  = {2},
  pages   = {163--185},
  year    = {1998},
}

@article{mackay1992,
  author  = {MacKay, David J. C.},
  title   = {Bayesian interpolation},
  journal = {Neural Comput.},
  volume  = {4},
  number  = {3},
  pages   = {415--447},
  year    = {1992},
}

@book{Jeffreys1961,
  author    = {Jeffreys, Harold},
  title     = {Theory of Probability},
  edition   = {3rd},
  publisher = {Oxford University Press},
  year      = {1961},
}

@article{rue2009,
  author  = {Rue, H{\aa}vard and Martino, Sara and Chopin, Nicolas},
  title   = {Approximate {B}ayesian inference for latent {G}aussian models
             by using integrated nested {L}aplace approximations},
  journal = {J. R. Stat. Soc. Ser. B},
  volume  = {71},
  number  = {2},
  pages   = {319--392},
  year    = {2009},
  doi     = {10.1111/j.1467-9868.2008.00700.x},
}

@article{gomezrubio2018,
  author        = {G{\'o}mez-Rubio, Virgilio and Rue, H{\aa}vard},
  title         = {{M}arkov chain {M}onte {C}arlo with the Integrated Nested {L}aplace Approximation},
  journal       = {Stat. Comput.},
  volume        = {28},
  pages         = {1033--1051},
  year          = {2018},
  eprint        = {1701.07844},
  archivePrefix = {arXiv},
}

@article{hadzhiyska2023,
  author        = {Hadzhiyska, Boryana and Wolz, Kai and Azzoni, Stefano and
                   Alonso, David and Garc{\'i}a-Garc{\'i}a, Carlos and others},
  title         = {Cosmology with 6 parameters in the Stage-{IV} era:
                   efficient marginalisation over nuisance parameters},
  journal       = {Open J. Astrophys.},
  year          = {2023},
  doi           = {10.21105/astro.2301.11895},
  eprint        = {2301.11895},
  archivePrefix = {arXiv},
}

@article{kristensen2016,
  author  = {Kristensen, Kasper and Nielsen, Anders and Berg, Casper W. and
             Skaug, Hans and Bell, Brad},
  title   = {{TMB}: Automatic Differentiation and {L}aplace Approximation},
  journal = {J. Stat. Softw.},
  volume  = {70},
  number  = {5},
  pages   = {1--21},
  year    = {2016},
  eprint  = {1509.00660},
  archivePrefix = {arXiv},
}

@inproceedings{simpson2021,
  author        = {Simpson, Fergus and Lalchand, Vidhi},
  title         = {Marginalised {G}aussian Processes with Nested Sampling},
  booktitle     = {Advances in Neural Information Processing Systems (NeurIPS)},
  year          = {2021},
  eprint        = {2010.16344},
  archivePrefix = {arXiv},
}

@inproceedings{lai2023,
  author        = {Lai, Justin and Burroni, Javier and Guan, Hui and Sheldon, Daniel},
  title         = {Automatically Marginalized {MCMC} in Probabilistic Programming},
  booktitle     = {Proceedings of the 40th International Conference on
                   Machine Learning (ICML)},
  year          = {2023},
  eprint        = {2302.00564},
  archivePrefix = {arXiv},
}

@book{rasmussen_williams_2006,
  author    = {Rasmussen, Carl Edward and Williams, Christopher K. I.},
  title     = {Gaussian Processes for Machine Learning},
  publisher = {MIT Press},
  year      = {2006},
}

@inproceedings{shah2014,
  author        = {Shah, Amar and Wilson, Andrew Gordon and Ghahramani, Zoubin},
  title         = {Student-$t$ Processes as Alternatives to {G}aussian Processes},
  booktitle     = {Proceedings of the 17th International Conference on
                   Artificial Intelligence and Statistics (AISTATS)},
  year          = {2014},
  eprint        = {1402.4306},
  archivePrefix = {arXiv},
}

@inproceedings{luu2024,
  author    = {Luu, Son and Xu, Zuheng and Surjanovic, Nikola and
               Biron-Lattes, Miguel and Campbell, Trevor and
               Bouchard-C{\^o}t{\'e}, Alexandre},
  title     = {Is {G}ibbs Sampling Faster than {H}amiltonian {M}onte {C}arlo on {GLM}s?},
  booktitle = {Proceedings of the 28th International Conference on
               Artificial Intelligence and Statistics (AISTATS)},
  year      = {2025},
}

@article{hutchinson1990,
  author  = {Hutchinson, M. F.},
  title   = {A stochastic estimator of the trace of the influence matrix
             for {L}aplacian smoothing splines},
  journal = {Commun. Stat. Simul. Comput.},
  volume  = {19},
  number  = {2},
  pages   = {433--450},
  year    = {1990},
}

@article{savchenko2025,
  author        = {Savchenko, Oleg and Abell{\'a}n, Guillermo Franco and List, Florian and
                   {Anau Montel}, Noemi and Weniger, Christoph},
  title         = {Fast Sampling of Cosmological Initial Conditions with {G}aussian Neural
                   Posterior Estimation},
  journal       = {arXiv preprint},
  year          = {2025},
  eprint        = {2502.03139},
  archivePrefix = {arXiv},
  primaryClass  = {astro-ph.CO},
}

@misc{jax2018,
  author  = {Bradbury, James and Frostig, Roy and Hawkins, Peter and
             Johnson, Matthew James and Leary, Chris and Maclaurin, Dougal and
             Necula, George and Paszke, Adam and {VanderPlas}, Jake and
             {Wanderman-Milne}, Skye and Zhang, Qiao},
  title   = {{JAX}: composable transformations of {Python+NumPy} programs},
  url     = {http://github.com/jax-ml/jax},
  year    = {2018},
}

@misc{Cabezas2024,
  author        = {Cabezas, Alberto and Corenflos, Adrien and Lao, Junpeng
                   and Louf, R{\'e}mi},
  title         = {{BlackJAX}: Composable {B}ayesian inference in {JAX}},
  year          = {2024},
  eprint        = {2402.10797},
  archivePrefix = {arXiv},
}

@misc{desdoigts2024,
  author        = {Desdoigts, Louis and Pope, Benjamin and Dennis, Jordan and
                   Tuthill, Peter},
  title         = {Differentiable Optics with d{L}ux {I}: Deep Calibration of
                   {F}lat {F}ield and {P}hase {R}etrieval with {A}utomatic
                   {D}ifferentiation},
  year          = {2024},
  eprint        = {2406.08703},
  archivePrefix = {arXiv},
  primaryClass  = {astro-ph.IM},
  doi           = {10.1117/1.JATIS.9.2.028007},
}

@article{brout2022,
  author        = {Brout, Dillon and Scolnic, Dan and Popovic, Brodie and Riess, Adam G. and Carr, Anthony and Zuntz, Joe and Kessler, Rick and Davis, Tamara M. and Hinshaw, Gary and Jones, David and others},
  title         = {The {Pantheon+} Analysis: Cosmological Constraints},
  journal       = {Astrophys. J.},
  volume        = {938},
  number        = {2},
  pages         = {110},
  year          = {2022},
  doi           = {10.3847/1538-4357/ac8e04},
  eprint        = {2202.04077},
  archivePrefix = {arXiv},
}

@article{tripp1998,
  author  = {Tripp, R.},
  title   = {A two-parameter luminosity correction for {Type~Ia} supernovae},
  journal = {Astron. Astrophys.},
  volume  = {331},
  pages   = {815--820},
  year    = {1998},
}

@article{Kim1998,
  author  = {Kim, Sangjoon and Shephard, Neil and Chib, Siddhartha},
  title   = {Stochastic Volatility: Likelihood Inference and Comparison with {ARCH} Models},
  journal = {Review of Economic Studies},
  volume  = {65},
  number  = {3},
  pages   = {361--393},
  year    = {1998},
}

@misc{Sountsov2020,
  author = {Sountsov, Pavel and Radul, Alexey and others},
  title  = {Inference Gym},
  year   = {2020},
  url    = {https://pypi.org/project/inference-gym/},
}

@article{Hahn2023,
  author        = {Hahn, Oliver and List, Florian and Porqueres, Natalia},
  title         = {{DISCO-DJ} {I}: a differentiable {E}instein--{B}oltzmann solver for cosmology},
  journal       = {arXiv preprint},
  year          = {2023},
  eprint        = {2311.03291},
  archivePrefix = {arXiv},
}

@article{Lewis2013,
  author        = {Lewis, Antony},
  title         = {Efficient sampling of fast and slow cosmological parameters},
  journal       = {Phys. Rev. D},
  volume        = {87},
  pages         = {103529},
  year          = {2013},
  eprint        = {1304.4473},
  archivePrefix = {arXiv},
}

@misc{McIntoshSmith2024,
  author        = {McIntosh-Smith, Simon and Alam, S. R. and Woods, Chris},
  title         = {{Isambard-AI}: a leadership class supercomputer optimised specifically for {A}rtificial {I}ntelligence},
  year          = {2024},
  eprint        = {2410.11199},
  archivePrefix = {arXiv},
  doi           = {10.48550/arXiv.2410.11199},
}

@book{GelmanBDA2013,
  author    = {Gelman, Andrew and Carlin, John B. and Stern, Hal S. and Dunson, David B. and Vehtari, Aki and Rubin, Donald B.},
  title     = {Bayesian Data Analysis},
  edition   = {3rd},
  publisher = {CRC Press},
  year      = {2013},
}

@article{Jasche2013,
  author        = {Jasche, Jens and Wandelt, Benjamin D.},
  title         = {Bayesian physical reconstruction of initial conditions from large-scale structure surveys},
  journal       = {Mon. Not. R. Astron. Soc.},
  volume        = {432},
  pages         = {894--913},
  year          = {2013},
  eprint        = {1203.3639},
  archivePrefix = {arXiv},
}

@book{CoverThomas2006,
  author    = {Cover, Thomas M. and Thomas, Joy A.},
  title     = {Elements of Information Theory},
  edition   = {2nd},
  publisher = {Wiley},
  year      = {2006},
}

@article{Blondel2022,
  author        = {Blondel, Mathieu and Berthet, Quentin and Cuturi, Marco and Frostig, Roy and Hoyer, Stephan and Llinares-L{\'o}pez, Felipe and Pedregosa, Fabian and Vert, Jean-Philippe},
  title         = {Efficient and modular implicit differentiation},
  journal       = {Advances in Neural Information Processing Systems},
  volume        = {35},
  year          = {2022},
  eprint        = {2105.15183},
  archivePrefix = {arXiv},
}

@article{McEwen2021,
  author        = {McEwen, Jason D. and Wallis, Christopher G. R. and Price, Matthew A. and Spurio Mancini, Alessio},
  title         = {Machine learning assisted {B}ayesian model comparison: learnt harmonic mean estimator},
  journal       = {arXiv preprint},
  year          = {2021},
  eprint        = {2111.12720},
  archivePrefix = {arXiv},
}

@article{Bevins2022a,
  author        = {Bevins, Harry T. J. and Handley, William J. and Lemos, Pablo and Sims, Peter H. and {de Lera Acedo}, Eloy and Fialkov, Anastasia and Alsing, Justin},
  title         = {Marginal post processing of {B}ayesian inference products with normalizing flows and kernel density estimators},
  journal       = {Mon. Not. R. Astron. Soc.},
  year          = {2023},
  eprint        = {2205.12841},
  archivePrefix = {arXiv},
}

@article{Bevins2022b,
  author        = {Bevins, Harry T. J. and Handley, Will and Lemos, Pablo and Sims, Peter H. and {de Lera Acedo}, Eloy and Fialkov, Anastasia},
  title         = {Marginal {B}ayesian statistics using masked autoregressive flows and kernel density estimators with examples in cosmology},
  journal       = {RAS Techniques and Instruments},
  year          = {2023},
  eprint        = {2207.11457},
  archivePrefix = {arXiv},
}

@article{Hu2024,
  author        = {Hu, Zixiao and Baryshnikov, Artem and Handley, Will},
  title         = {aeons: approximating the end of nested sampling},
  journal       = {Mon. Not. R. Astron. Soc.},
  year          = {2024},
  eprint        = {2312.00294},
  archivePrefix = {arXiv},
}

@article{LiuNocedal1989,
  author  = {Liu, Dong C. and Nocedal, Jorge},
  title   = {On the limited memory {BFGS} method for large scale optimization},
  journal = {Math. Program.},
  volume  = {45},
  pages   = {503--528},
  year    = {1989},
  doi     = {10.1007/BF01589116},
}

@article{Zhang2022pathfinder,
  author        = {Zhang, Lu and Carpenter, Bob and Gelman, Andrew and Vehtari, Aki},
  title         = {Pathfinder: Parallel quasi-{N}ewton variational inference},
  journal       = {J. Mach. Learn. Res.},
  volume        = {23},
  number        = {306},
  pages         = {1--49},
  year          = {2022},
  eprint        = {2108.03782},
  archivePrefix = {arXiv},
}

@article{Papaspiliopoulos2007,
  author  = {Papaspiliopoulos, Omiros and Roberts, Gareth O. and Sk{\"o}ld, Martin},
  title   = {A general framework for the parametrization of hierarchical models},
  journal = {Stat. Sci.},
  volume  = {22},
  number  = {1},
  pages   = {59--73},
  year    = {2007},
  doi     = {10.1214/088342307000000014},
}

@incollection{Betancourt2015,
  author        = {Betancourt, Michael and Girolami, Mark},
  title         = {Hamiltonian {M}onte {C}arlo for hierarchical models},
  booktitle     = {Current Trends in {B}ayesian Methodology with Applications},
  publisher     = {CRC Press},
  year          = {2015},
  eprint        = {1312.0906},
  archivePrefix = {arXiv},
}

@article{kroupa2025resonances,
  title={Resonances in reflective Hamiltonian Monte Carlo},
  author={Kroupa, Namu and Cs{\'a}nyi, G{\'a}bor and Handley, Will},
  journal={Physical Review E},
  volume={111},
  number={4},
  pages={045308},
  year={2025},
  publisher={APS}
}

@misc{beskos2010optimaltuninghybridmontecarlo,
      title={Optimal tuning of the Hybrid Monte-Carlo Algorithm}, 
      author={Alexandros Beskos and Natesh S. Pillai and Gareth O. Roberts and Jesus M. Sanz-Serna and Andrew M. Stuart},
      year={2010},
      eprint={1001.4460},
      archivePrefix={arXiv},
      primaryClass={math.PR},
      url={https://arxiv.org/abs/1001.4460}, 
}

@article{Trotta_2008,
   title={Bayes in the sky: Bayesian inference and model selection in cosmology},
   volume={49},
   ISSN={1366-5812},
   url={http://dx.doi.org/10.1080/00107510802066753},
   DOI={10.1080/00107510802066753},
   number={2},
   journal={Contemporary Physics},
   publisher={Informa UK Limited},
   author={Trotta, Roberto},
   year={2008},
   month=mar, pages={71–104} }

@article{mlreview,
  author  = {Llorente, F. and Martino, L. and Delgado, D. and L\'{o}pez-Santiago, J.},
  journal = {SIAM Review},
  number  = {1},
  pages   = {3-58},
  title   = {Marginal Likelihood Computation for Model Selection and Hypothesis Testing: An Extensive Review},
  volume  = {65},
  year    = {2023}
}

@inproceedings{petrosyan2023supernest,
  title={SuperNest: accelerated nested sampling applied to astrophysics and cosmology},
  author={Petrosyan, Aleksandr and Handley, Will},
  booktitle={Physical Sciences Forum},
  volume={5},
  number={1},
  pages={51},
  year={2023},
  organization={MDPI}
}

@InProceedings{pmlr-v119-wenzel20a,
  title = 	 {How Good is the {B}ayes Posterior in Deep Neural Networks Really?},
  author =       {Wenzel, Florian and Roth, Kevin and Veeling, Bastiaan and Swiatkowski, Jakub and Tran, Linh and Mandt, Stephan and Snoek, Jasper and Salimans, Tim and Jenatton, Rodolphe and Nowozin, Sebastian},
  booktitle = 	 {Proceedings of the 37th International Conference on Machine Learning},
  pages = 	 {10248--10259},
  year = 	 {2020},
  editor = 	 {III, Hal Daumé and Singh, Aarti},
  volume = 	 {119},
  series = 	 {Proceedings of Machine Learning Research},
  month = 	 {13--18 Jul},
  publisher =    {PMLR},
  url = 	 {https://proceedings.mlr.press/v119/wenzel20a.html}
}

@inproceedings{pmlr-v130-hoffman21a,
  title     = { An Adaptive-MCMC Scheme for Setting Trajectory Lengths in Hamiltonian Monte Carlo },
  author    = {Hoffman, Matthew and Radul, Alexey and Sountsov, Pavel},
  booktitle = {Proceedings of The 24th International Conference on Artificial Intelligence and Statistics},
  pages     = {3907--3915},
  year      = {2021},
  editor    = {Banerjee, Arindam and Fukumizu, Kenji},
  volume    = {130},
  series    = {Proceedings of Machine Learning Research},
  month     = {13--15 Apr},
  publisher = {PMLR},
  url       = {https://proceedings.mlr.press/v130/hoffman21a.html}
}

@misc{rioudurand2023adaptivetuningmetropolisadjusted,
  title         = {Adaptive Tuning for Metropolis Adjusted Langevin Trajectories},
  author        = {Lionel Riou-Durand and Pavel Sountsov and Jure Vogrinc and Charles C. Margossian and Sam Power},
  year          = {2023},
  eprint        = {2210.12200},
  archiveprefix = {arXiv},
  primaryclass  = {stat.CO},
  url           = {https://arxiv.org/abs/2210.12200}
}

@misc{gorinova2019automaticreparameterisationprobabilisticprograms,
  title         = {Automatic Reparameterisation of Probabilistic Programs},
  author        = {Maria I. Gorinova and Dave Moore and Matthew D. Hoffman},
  year          = {2019},
  eprint        = {1906.03028},
  archiveprefix = {arXiv},
  primaryclass  = {stat.ML},
  url           = {https://arxiv.org/abs/1906.03028}
}

@article{Millea:2021had,
    author = "Millea, Marius and Seljak, Uros",
    title = "{Marginal unbiased score expansion and application to CMB lensing}",
    eprint = "2112.09354",
    archivePrefix = "arXiv",
    primaryClass = "astro-ph.CO",
    doi = "10.1103/PhysRevD.105.103531",
    journal = "Phys. Rev. D",
    volume = "105",
    number = "10",
    pages = "103531",
    year = "2022"
}

@article{popcosmos,
author = {Alsing, Justin and Thorp, Stephen and Deger, Sinan and Peiris, Hiranya and Leistedt, Boris and Mortlock, Daniel and Leja, Joel},
year = {2024},
month = {09},
pages = {12},
title = {pop-cosmos: A Comprehensive Picture of the Galaxy Population from COSMOS Data},
volume = {274},
journal = {The Astrophysical Journal Supplement Series},
doi = {10.3847/1538-4365/ad5c69}
}

@article{BAYESN,
    author = {Mandel, Kaisey S and Thorp, Stephen and Narayan, Gautham and Friedman, Andrew S and Avelino, Arturo},
    title = {A hierarchical Bayesian SED model for Type Ia supernovae in the optical to near-infrared},
    journal = {Monthly Notices of the Royal Astronomical Society},
    volume = {510},
    number = {3},
    pages = {3939-3966},
    year = {2021},
    month = {12},
    issn = {0035-8711},
    doi = {10.1093/mnras/stab3496},
    url = {https://doi.org/10.1093/mnras/stab3496},
    eprint = {https://academic.oup.com/mnras/article-pdf/510/3/3939/42182470/stab3496.pdf},
}
\bibliographystyle{plainnat}


\appendix

\section{Related Work}
\label{sec:related}

\paragraph{Type II maximum likelihood and the evidence approximation.}
\citet{mackay1992} introduced the \emph{evidence approximation} (Type~II maximum likelihood), which applies the Laplace approximation to obtain a closed-form surrogate for the marginal likelihood and then \emph{maximises} it over hyperparameters $\vtheta$, yielding a point estimate.
This is the standard approach for kernel hyperparameter selection in Gaussian process classification~\citep{rasmussen_williams_2006}, where the Laplace approximation over the latent function values $\vz$ is combined with gradient-based optimisation of $(\log\ell, \log A)$.
The automatic differentiation package TMB~\citep{kristensen2016} generalises this pattern to arbitrary hierarchical models with random effects, using autodiff to compute the Laplace-approximated marginal likelihood and then optimising over $\vtheta$.
Crucially, TMB's $\vtheta$ estimate is \emph{not} the joint MAP of $(\vtheta, \vz)$: the $-\tfrac{1}{2}\log\det\mH(\vtheta)$ term in $\mathcal{L}_\mathrm{ALCS}(\vtheta)$ acts as an Occam penalty that accounts for the volume of the latent posterior at each $\vtheta$, so TMB finds the MAP of the \emph{marginalised} posterior $p(\vtheta \mid \data) \propto \mathcal{L}_\mathrm{ALCS}(\vtheta)\,\pi(\vtheta)$.
Stan's \texttt{laplace\_marginals}~\citep{stan2023} implements the same inner step (Laplace approximation of $\vz$ at fixed $\vtheta$) and can similarly be combined with outer optimisation over $\vtheta$.
ALCS differs from all of these in a crucial respect: rather than \emph{optimising} $\vtheta$, it \emph{integrates} over $\vtheta$ using nested sampling, performing a proper Bayesian marginalisation and directly computing the evidence $\mathcal{Z}$.
We demonstrate in \cref{sec:experiments} that this distinction matters even when both methods use the same Laplace approximation over $\vz$.

\paragraph{Laplace marginalisation inside sampling: INLA and successors.}
The idea of fixing $\vtheta$ on a grid and applying the Laplace approximation to marginalise $\vz$ analytically was systematised by \citet{rue2009} as the Integrated Nested Laplace Approximation (INLA), designed for latent Gaussian models.
INLA targets posterior marginals $p(\theta_j \mid \data)$ rather than the evidence, and its deterministic grid over $\vtheta$ limits practical application to $d_\theta \lesssim 5$.
\citet{gomezrubio2018} extended INLA by replacing the grid with Metropolis--Hastings sampling over $\vtheta$, allowing higher-dimensional $\vtheta$ but still targeting posteriors rather than evidence.
A more accurate alternative to the Laplace approximation for the inner $\vz$ step is Expectation Propagation (EP;~\citealt{minka2001}), which fits a product of Gaussian site approximations rather than a single Gaussian at the mode, and is standard in GP classification libraries~\citep{rasmussen_williams_2006}.
EP can better capture asymmetric or non-Gaussian factors, but it does not yield a closed-form expression analogous to \cref{eq:ALCS}, and differentiating through the EP fixed-point equations to optimise or sample over $\vtheta$ requires implicit differentiation, making it substantially harder to compose with an outer sampler.
\citet{hadzhiyska2023} applied the same Laplace-over-$\vz$ marginalisation to Stage~IV weak lensing, using HMC over $\vtheta$ and hand-derived model-specific Hessians.
ALCS takes this programme to its natural conclusion in two ways.
First, ALCS uses JAX automatic differentiation for Hessians, requiring only a differentiable log-posterior rather than analytic derivation.
Second, and more fundamentally, ALCS uses nested sampling over $\vtheta$ rather than MCMC or HMC.
Once Laplace marginalisation has reduced the inference problem to $d_\theta \sim 10$ dimensions, nested sampling is the natural choice: it directly produces the Bayesian evidence $\mathcal{Z}$ with uncertainty estimates, handles multimodal $\vtheta$ posteriors without modification, and requires no post-processing such as bridge sampling~\citep{gelman_simulating_1998} to recover $\mathcal{Z}$ from posterior samples.

\paragraph{Exact marginalisation with nested sampling.}
\citet{simpson2021} showed that for GP models with Gaussian likelihoods, where the marginalisation over $\vz$ is \emph{exact}, running nested sampling over $\vtheta$ yields substantially more accurate evidence estimates and better-calibrated hyperparameter posteriors than Type~II ML.
Their work directly establishes NS as the right $\vtheta$-sampler for this problem class and motivates ALCS's design choices.
ALCS extends \citet{simpson2021} from the exact-GP setting to the general non-conjugate case: the Laplace approximation provides a tractable, differentiable surrogate for the intractable marginal likelihood $\mathcal{L}_\mathrm{marg}(\vtheta)$ whenever the latent posterior $p(\vz \mid \vtheta, \data)$ is approximately Gaussian.

\paragraph{Automatic marginalisation in probabilistic programming.}
\citet{lai2023} propose automating analytic marginalisation of conjugate local variables inside HMC in probabilistic programming, giving efficiency gains analogous to ALCS's elimination of $\vz$ from the NS problem.
Their work exploits exact conjugate marginalisation rather than the Laplace approximation, and targets posteriors rather than evidence.
The growing interest in analytic marginalisation during sampling, across INLA, \citet{gomezrubio2018}, \citet{lai2023}, and ALCS, reflects a broader recognition that removing tractable dimensions from a sampler improves both efficiency and the quality of the resulting inference.

\paragraph{Nested sampling with structured likelihoods.}
The concurrent NS-SwiG algorithm~\citep{NS-SwiG} targets the same class of hierarchical models from a complementary angle: rather than marginalising out the latents, it exploits the factorised likelihood to reduce the cost of each nested sampling step from $\mathcal{O}(J^2)$ to $\mathcal{O}(J)$ by caching per-group likelihood contributions.
NS-SwiG retains the full joint NS treatment of both $\vtheta$ and $\vz$, at the cost of a quadratic rather than cubic dimension scaling.
ALCS achieves more aggressive dimension reduction by removing $\vz$ from the NS problem entirely, at the cost of the Gaussian approximation over $\vz$.
The two methods are complementary: ALCS is preferable when $\vz \mid \vtheta, \data$ is approximately Gaussian and $d_z \gg d_\theta$; NS-SwiG is preferable when the exact $\vz$ posterior is required.
Proximal nested sampling~\citep{cai2022} addresses a different regime of high-dimensional NS, using Moreau envelopes to handle non-smooth log-concave likelihoods in imaging; it is complementary to ALCS in that it does not rely on a Gaussian approximation.

\paragraph{Scalable posterior sampling.}
For problems where evidence is not required, \citet{luu2024} show that a Compute-Graph Gibbs sampler exploiting the computational structure of GLM likelihoods reduces per-sweep cost from $\mathcal{O}(d^2)$ to $\mathcal{O}(d)$, outperforming HMC for high-dimensional GLMs.
This is purely a posterior sampling method and does not address evidence computation.

\paragraph{Structured precision matrices.}
For latent spaces of very high dimension ($d_z \sim 10^6$, e.g.\ field-level cosmological inference) even a single Hessian evaluation is prohibitive.
A concrete example is $\partial$Lux~\citep{desdoigts2024}, a JAX differentiable physical optics pipeline in which the full Hessian of the instrument forward model is intractable, preventing direct Laplace marginalisation over the high-dimensional pixel-level latents.
\citet{savchenko2025} develop techniques for Hessian-vector products and structured precision matrix computation at field-level scales, using the Hutchinson stochastic trace estimator~\citep{hutchinson1990} for log-determinant computation.
In principle, such methods could serve as the inner Hessian step of ALCS, extending it to field-level latent spaces, an avenue we leave to future work.

\section{Discussion}
\label{app:discussion}

\paragraph{Prior work.}
The idea of marginalising nuisance parameters is central to Bayesian inference and predates numerical methods: conjugate prior constructions are an exact analytic instance of this principle.
In the numerical setting, \citet{hadzhiyska2023} applied Laplace marginalisation to Stage IV weak lensing with hand-derived Hessians and HMC for the remaining parameters; \citet{Bevins2022a,Bevins2022b} learned marginal posteriors from existing posterior samples via normalising flows.
To our knowledge, ALCS is the first framework to combine an \emph{automatic} variant of Laplace marginalisation (requiring only a differentiable log-posterior with no hand-derived derivatives) with a nested sampling outer particle loop.

\paragraph{Limitations.}
\emph{Unimodality of the latent conditional.}
ALCS requires $p(\vz \mid \vtheta, \data)$ to be approximately unimodal at each $\vtheta$ in the posterior support.
If the latent conditional is multimodal, the MAP optimisation may converge to the wrong mode and the resulting evidence estimate is incorrect with no obvious warning sign.
This is a fundamental limitation shared by all methods that scale to high latent dimension (see \cref{sec:method}), but it means ALCS should not be applied without prior knowledge that the latent posterior is well-behaved.

\emph{Gaussian tail approximation.}
Even when the latent conditional is unimodal, the Laplace approximation can misestimate the evidence when the posterior tails deviate significantly from Gaussian.
The Student-$t$ extension of \cref{subsec:student_t} partially addresses the heavy-tailed case, but the correction degrades when the Gaussian likelihood progressively suppresses the heavy tails as $N$ grows.
The canonical failure case illustrated here is the tanh funnel (\cref{subsec:funnel}): the latent non-Gaussianity varies strongly with $\vtheta$, so no single Gaussian approximation is adequate across the full hyperparameter space.

\emph{Hessian scaling.}
The dense Hessian requires $\mathcal{O}(d_z^2)$ memory and $\mathcal{O}(d_z \times T_\mathcal{L})$ compute per ALCS call.
Structured sparsity (\cref{app:hessian}) reduces this substantially for models with known Hessian structure, but at field-level scales ($d_z \sim 10^6$) even a single Hessian-vector product may be prohibitive and stochastic trace estimation~\citep{savchenko2025} would be needed.

\emph{The Laplace assumption cannot be verified from the NS output alone.}
The NS run yields $\vtheta$ samples and an evidence estimate, but no direct window into whether $p(\vz \mid \vtheta, \data)$ was well-approximated at each $\vtheta$.
The Hessian-preconditioned gradient norm $\nabla_{\vz}^\top \mH^{-1} \nabla_{\vz}\log p^*(\hat{\vz})$ at the MAP confirms optimiser convergence but not Gaussian fidelity.
A post-hoc diagnostic is possible: for each $\vtheta$ sample, draw $\vz \sim \mathcal{N}(\hat{\vz}(\vtheta), \mH(\vtheta)^{-1})$ and importance weight against the true $p(\vz \mid \vtheta, \data)$, monitoring the effective sample size (ESS).
Low ESS at particular $\vtheta$ values flags where the approximation breaks down without requiring a full joint NS run.
We provide an illustrative application of this diagnostic in \cref{app:is_diagnostic}.

\emph{Optimiser convergence.}
L-BFGS convergence is not guaranteed for poorly conditioned latent spaces.
The preconditioned gradient norm (above) should be monitored; a large residual invalidates the Hessian computation and the returned $\mathcal{L}_\mathrm{ALCS}(\vtheta)$.

\paragraph{Extensions.}
\emph{Richer local approximations.}
The Gaussian Laplace approximation is the simplest member of a broader family of local approximations: fit a tractable distribution at the mode by matching derivatives via autodiff, then integrate analytically.
The Student-$t$ extension of \cref{subsec:student_t} matches the second and fourth derivatives; a skewed approximation such as a $\chi^2$ or log-normal could additionally match the third derivative, capturing asymmetric latent posteriors.
More generally, any distribution with a known normalisation constant and derivatives that can be matched to those of the log-posterior at the mode is a candidate.
Whether such extensions lead to reliably better evidence estimates in practice, rather than introducing new failure modes from imperfect moment matching, is an open question, but the autodiff infrastructure developed here makes them straightforward to explore.

\emph{Field-level inference.}
The structured Hessian machinery of \cref{app:hessian} is a step towards very large latent spaces, but at field-level scales ($d_z \sim 10^6$) even computing three HVPs per ALCS call may be expensive.
Combining stochastic Hessian-vector products~\citep{savchenko2025} with the Hutchinson trace estimator for the log-determinant would extend ALCS to this regime, at the cost of introducing stochasticity into the marginalised likelihood.

\emph{Post-hoc assumption checking.}
The IS/ESS diagnostic of \cref{app:is_diagnostic} is currently presented as a verification tool, but it could be used more actively: regions of low ESS could trigger a local refinement step (for example, a short MCMC chain at that $\vtheta$ to obtain a better evidence estimate) while leaving the bulk of the NS run unaffected. The viability and effectiveness of this would be highly problem-dependent, varying with latent dimension and structure.

\section{Implementation Details}
\label{app:implementation}

\paragraph{JAX and automatic differentiation.}
The optimisation and Hessian in \cref{eq:map,eq:hessian} require only a differentiable log-posterior; no analytic derivation of gradients or Hessians is needed.
We implement ALCS in JAX~\citep{jax2018}, using \texttt{jax.hessian} (implemented as forward-over-reverse automatic differentiation) for the Hessian and L-BFGS for the optimisation.
The collapsed marginal likelihood $\mathcal{L}_\mathrm{ALCS}(\vtheta)$ is a drop-in replacement for any outer sampler; in the experiments of \cref{sec:experiments} we use the BlackJAX~\citep{Cabezas2024} nested slice sampling implementation.
This makes ALCS applicable to any model whose log-posterior is expressible as a JAX function.

\paragraph{Prior whitening.}
Optimisation over $\vz$ is greatly improved by working in a whitened coordinate system. Even though L-BFGS builds an internal approximation of the inverse Hessian, we find prior whitening to be practically essential for high-dimensional latent spaces. Whitening ensures the initial optimization steps are well-scaled before the L-BFGS history buffer is fully populated, and it restricts the algorithm's limited-memory curvature updates to learning only the non-linearities introduced by the likelihood, dramatically improving convergence stability. Given an initial estimate $H_0$ of the prior precision (e.g.\ the prior precision matrix), we transform $\tilde{\vz} = H_0^{1/2}(\vz - \vz_0)$ so that the prior is approximately unit Gaussian.
The Hessian in the original coordinates is recovered as $\mH = H_0^{1/2}\,\tilde{\mH}\,H_0^{1/2}$. 

\paragraph{Warm-start preconditioning.}
For problems where the Hessian varies slowly across $\vtheta$, a substantial speedup is available.
We perform a single full optimisation and Hessian computation at a fiducial $\vtheta_\mathrm{fid}$ to obtain $\hat{\vz}_\mathrm{fid}$ and $H_\mathrm{fid}$, then use these as the initial point and whitening transform for all subsequent likelihood evaluations.
Crucially, this does \emph{not} bias the evidence: the final Hessian in \cref{eq:ALCS} is computed at the actual MAP $\hat{\vz}(\vtheta)$, not the fiducial approximation.
The warm start only affects the optimisation path, not the final answer.

\paragraph{Caching $\vtheta$-dependent computations.}
In many physical models the log-likelihood decomposes into a $\vtheta$-dependent part that is expensive to evaluate (e.g.\ a power spectrum or transfer function computed by an Einstein--Boltzmann solver~\citep{Hahn2023}) and a $\vz$-dependent part that, given those outputs, is cheap.
This separation is the cosmological analogue of the fast--slow parameter structure exploited in MCMC~\citep{Lewis2013}.
In the ALCS inner loop, $\vtheta$ is fixed while L-BFGS optimises over $\vz$: the expensive $\vtheta$-dependent outputs can therefore be computed once before the optimisation begins and cached for the duration, so that each gradient evaluation in \texttt{jax.grad} accesses only the cheap $\vz$-dependent computation.
This requires no change to the ALCS interface; it is a standard application of JAX's \texttt{jax.lax.stop\_gradient} or equivalent pre-computation outside the differentiated function.

\paragraph{Block-diagonal factorisation.}
For hierarchical models where the log-likelihood factorises over $J$ independent groups,
\begin{equation}
\log \mathcal{L}(\data \mid \vtheta, \vz) = \sum_{j=1}^J \log \mathcal{L}_j(\data_j \mid \vtheta, \vz_j),
\label{eq:factorised_app}
\end{equation}
the Hessian over $\vz = (\vz_1, \ldots, \vz_J)$ is block-diagonal.
The log-determinant then decomposes as $\log\det\mH = \sum_j \log\det\mH_j$, and the per-group optimisations and Hessian computations are independent.
We exploit this with \texttt{jax.vmap} to parallelise all $J$ groups simultaneously on GPU, recovering the full Hessian at cost $\mathcal{O}(d_{z,\mathrm{block}})$ per group rather than $\mathcal{O}(d_z)$. 
We also extend this to banded or tridiagonal Hessians (nearest-neighbour covariance only) in \Cref{app:hessian}.

\section{Derivation of the Student-$t$ Laplace Formula}
\label{app:derivation}

\subsection{Setup and parameterisation}

Following \citet{shah2014}, a multivariate Student-$t$ with $\nu$ degrees of
freedom, location $\bm{\mu}$, and covariance matrix $\bm{K}$ has unnormalised
kernel
\begin{equation}
  p^*(\vz) \;\propto\;
  \left(1 + \frac{(\vz-\bm{\mu})^\top \bm{K}^{-1}(\vz-\bm{\mu})}{\nu-2}\right)^{-(\nu+d_z)/2},
  \label{eq:mvt-kernel}
\end{equation}
with normalisation constant
\begin{equation}
  C_t
  = \frac{\Gamma\!\bigl((\nu+d_z)/2\bigr)}
         {\Gamma(\nu/2)\,\bigl[(\nu-2)\pi\bigr]^{d_z/2}\,|\bm{K}|^{1/2}}.
  \label{eq:mvt-norm}
\end{equation}
The parameterisation using $(\nu-2)$ in the denominator ensures that
$\mathrm{Cov}[\vz] = \bm{K}$ (the covariance equals the scale matrix), and
the distribution is well-defined for $\nu > 2$.
The Hessian of $-\log p^*$ at the mode $\bm{\mu}$ is
\begin{equation}
  \mH = -\nabla^2_{\vz}\log p^*\big|_{\vz=\bm{\mu}}
  = \frac{\nu+d_z}{(\nu-2)}\,\bm{K}^{-1}.
  \label{eq:mvt-hessian}
\end{equation}

\subsection{General Laplace formula}

For any normalised approximating density $q(\vz)$ that fits the mode and
curvature of $p^*$, the Laplace approximation to the evidence is
\begin{equation}
  \mathcal{Z}
  \;\approx\;
  \frac{p^*(\hat{\vz})}{q(\hat{\vz})},
  \qquad
  \log\mathcal{Z}
  \;\approx\;
  \log p^*(\hat{\vz}) - \log q(\hat{\vz}),
  \label{eq:laplace-general}
\end{equation}
where $\hat{\vz}$ is the MAP.
For a Gaussian $q$ with precision $\mH$, $\log q(\hat{\vz}) =
-\tfrac{d_z}{2}\log(2\pi)+\tfrac{1}{2}\log\det\mH$, recovering
\cref{eq:log_laplace}.

\subsection{Hessian-matched Student-$t$ in the whitened basis}

We wish to use a Student-$t$ approximating distribution that (i) has its mode
at $\hat{\vz}$ and (ii) matches the Hessian $\mH$ at the mode.
From \cref{eq:mvt-hessian}, condition (ii) requires
$\bm{K} = \tfrac{\nu+d_z}{\nu-2}\,\mH^{-1}$.

Working in the Cholesky-whitened basis $\bm{w} = L(\vz - \hat{\vz})$, where
$\mH = LL^\top$, reduces the problem to $d_z$ independent univariate problems.
In whitened coordinates the Hessian is the identity, so condition (ii) for each
dimension $j$ requires covariance $K_j = (\nu_j+1)/(\nu_j-2)$ (using $d_z = 1$
per dimension).
The change of variables contributes a Jacobian $|\det L|^{-1} =
|\mH|^{-1/2}$, so
\begin{equation}
  \log\mathcal{Z}
  = \log p^*(\hat{\vz}) - \tfrac{1}{2}\log\det\mH
  - \sum_{j=1}^{d_z} \log q_j(0),
  \label{eq:st-full-derive}
\end{equation}
where $\log q_j(0)$ is the log of the normalised 1D Student-$t$ PDF evaluated
at its mode.

\subsection{Per-dimension log-normalisation}

The normalised 1D Student-$t$ with covariance $K_j = (\nu_j+1)/(\nu_j-2)$ and
$\nu_j$ degrees of freedom has PDF
\begin{equation}
  q_j(w)
  = \frac{\Gamma\!\bigl((\nu_j+1)/2\bigr)}
         {\sqrt{\pi(\nu_j-2)K_j}\;\Gamma(\nu_j/2)}
    \left(1 + \frac{w^2}{(\nu_j-2)K_j}\right)^{-(\nu_j+1)/2}.
\end{equation}
Substituting $K_j = (\nu_j+1)/(\nu_j-2)$ so that $(\nu_j-2)K_j = \nu_j+1$:
\begin{equation}
  \log q_j(0)
  = \log\Gamma\!\left(\tfrac{\nu_j+1}{2}\right)
  - \log\Gamma\!\left(\tfrac{\nu_j}{2}\right)
  - \tfrac{1}{2}\log\!\bigl(\pi(\nu_j+1)\bigr).
  \label{eq:logq0}
\end{equation}
Substituting into \cref{eq:st-full-derive} and comparing with \cref{eq:ALCS}
gives the complete formula \cref{eq:student_full}.

\paragraph{Sanity check: Gaussian limit.}
As $\nu_j \to \infty$, Stirling's approximation gives
$\log\Gamma((\nu_j+1)/2) - \log\Gamma(\nu_j/2) \to \tfrac{1}{2}\log(\nu_j/2)$,
so $\log q_j(0) \to \tfrac{1}{2}\log(\nu_j/2) - \tfrac{1}{2}\log(\pi\nu_j)
= -\tfrac{1}{2}\log(2\pi)$.
Substituting into \cref{eq:st-full-derive}:
$\log\mathcal{Z} \to \log p^*(\hat{\vz}) - \tfrac{1}{2}\log\det\mH
+ \tfrac{d_z}{2}\log(2\pi)$, which is \cref{eq:log_laplace}. $\checkmark$

\paragraph{Equivalence with Shah et al.\ Corollary B.}
When all dimensions share the same $\nu$, \cref{eq:st-full-derive,eq:logq0}
combine to give
$\log\mathcal{Z} = \log p^*(\hat{\vz}) - \tfrac{1}{2}\log\det\mH
- \log\Gamma((\nu+d_z)/2) + \log\Gamma(\nu/2)
+ \tfrac{d_z}{2}\log(\pi(\nu+d_z))$,
which is exactly the formula obtained by applying \cref{eq:laplace-general}
with the full multivariate Student-$t$ normalisation \cref{eq:mvt-norm} and
$\bm{K} = \tfrac{\nu+d_z}{\nu-2}\mH^{-1}$.
The per-dimension parameterisation with independent $\nu_j$ extends this to the
heterogeneous case where non-Gaussianity varies across whitened directions.

\subsection{Estimating $\nu_j$}

A Student-$t$ distribution with $\nu$ degrees of freedom has excess kurtosis
$\kappa = 6/(\nu-4)$ for $\nu > 4$.
Given an estimate $\hat\kappa_j$ of the excess kurtosis of the marginal
posterior in whitened direction $j$, the degrees of freedom estimator is
\begin{equation}
  \hat\nu_j = 4 + \frac{6}{\hat\kappa_j}.
  \label{eq:nu-estimator}
\end{equation}
The kurtosis $\hat\kappa_j$ is estimated from the fourth derivative of the
log-posterior at the MAP (see \cref{eq:kurtosis-4d}), a purely local
computation that requires no additional sampling.
This estimator is exact when the conditional posterior in direction $j$ is
itself a Student-$t$, but can fail when the posterior shape is more complex
(e.g.\ the tanh funnel, \Cref{subsec:funnel}).
An alternative route to $\hat\kappa_j$ is importance sampling from the
Laplace Gaussian, but in this work we prefer to keep $\hat\nu_j$ estimation
purely local and free of additional stochastic calls; the fourth-derivative
estimator achieves this at negligible extra cost.
In practice we find the kurtosis estimator \cref{eq:nu-estimator} gives
$\hat\nu_j \approx 20$--$50$ for the Student-$t$ prior experiment
(\Cref{subsec:student_t_exp}).
This is the correct effective tail weight of the latent conditional posterior
$p(z_i \mid \vtheta, \data)$, which is lighter-tailed than the $\nu{=}5$
prior because the Gaussian likelihood contributes a quadratic (i.e.\ Gaussian)
factor that softens the tails.

\section{Experiment Details}
\label{app:exp_details}

\subsection{Supernova Cosmology}
\label{app:exp_sne}

\paragraph{Model specification.}
The Tripp formula~\citep{tripp1998} gives the distance modulus for supernova $i$ as
$\mu_i = m_{B,i} - M + \alpha\,x_{1,i} - \beta\,c_i$,
where $m_{B,i}$ is the peak apparent magnitude, $x_{1,i}$ is the light-curve stretch, $c_i$ is the colour, and $(M, \alpha, \beta)$ are population parameters absorbed into the hyperparameters $\vtheta$.
Each supernova has a Gaussian likelihood
$y_i \mid \vz_i, \vtheta \sim \mathcal{N}(\mu_i(\vtheta, \vz_i),\, \sigma_\text{obs}^2)$
and Gaussian conditional prior $\vz_i = (x_{1,i}, c_i) \mid \vtheta \sim \mathcal{N}(\bm{0}, \Sigma_\text{pr})$.
Because both factors are Gaussian, the marginalisation over $\vz_i$ is analytic, providing the exact reference $\log\mathcal{Z}_\text{true}$ used throughout \Cref{subsec:sne}.

\paragraph{Cosmological models.}
Two models are compared: $\Lambda$CDM ($d_\theta=2$, parameters $\Omega_m$ and $w_0=-1$ fixed) and $w$CDM ($d_\theta=3$, additionally varying $w_0$ and $w_a$).
Both models are run on each dataset; the Bayes factor $\ln(\mathcal{Z}_\Lambda/\mathcal{Z}_w)$ is a byproduct.

\paragraph{Data generation and Hessian structure.}
Synthetic datasets are generated at seed 42 with true cosmology $\Omega_m = 0.3$, redshifts uniformly spaced in $z_\text{obs} \in [0.1, 1.0]$.
Test 1 fixes $d_{z,\text{block}} = 2$ and varies $N \in \{64, 128, 256, 512, 1024, 2048\}$.
Test 2 fixes $N = 100$ and varies $d_{z,\text{block}} \in \{2, 4, 8, 16, 32, 64, 128, 256\}$ by adding synthetic photometric bands.
The likelihood factorises over objects, so the Hessian is block-diagonal with $N$ independent $d_{z,\text{block}} \times d_{z,\text{block}}$ blocks computed in parallel via \texttt{jax.vmap}.

\subsection{Student-$t$ Prior}
\label{app:exp_student}

\paragraph{Data generation.}
Data are generated at seed 42 with $\mu_\text{true} = 0$, $\sigma_\text{true} = 1$, $\sigma_\text{obs} = 1$, for $N_\text{obj} \in \{10, 20, 50, 100, 150\}$.
The hyperparameters have broad uniform priors: $\mu \in [-3, 3]$ and $\log\sigma \in [\log 0.1, \log 5]$.
Ground truth is obtained from a joint NS over $(\mu, \log\sigma, z_1, \ldots, z_{N_\text{obj}})$.

\paragraph{Effective tail weight.}
The true prior on each $z_i$ is Student-$t(\nu=5)$, but the Gaussian likelihood $y_i \mid z_i \sim \mathcal{N}(z_i, 1)$ contributes a quadratic term that softens the tails.
The conditional posterior $p(z_i \mid y_i, \vtheta)$ is therefore heavier-tailed than Gaussian but lighter-tailed than the $\nu=5$ prior.
The fourth-derivative estimator \cref{eq:nu-estimator} correctly captures this intermediate tail weight, returning $\hat\nu \approx 20$--$50$.

\subsection{Tanh Funnel}
\label{app:exp_funnel}

\paragraph{Data generation.}
A single dataset of $J=10$ observations is generated at seed 42 with $\theta_\text{true} = 0$ using the model of \cref{eq:tanh_funnel}.
The ground truth $\mathcal{L}_\text{true}(\theta)$ is computed by one-dimensional numerical quadrature (\texttt{scipy.integrate.quad}) at each $\theta$ value.

\paragraph{NS and grid configuration.}
Both the full joint NS and the ALCS outer NS use $m=500$ live points and terminate when $\log\mathcal{Z}_\text{live} - \log\hat{\mathcal{Z}} < -3$.
The marginalised log-likelihood comparison (\Cref{fig:funnel}a) evaluates $\log\mathcal{L}_\text{ALCS}(\theta)$ and $\log\mathcal{L}_\text{true}(\theta)$ on 60 uniformly-spaced $\theta$ values in $[-3, 4]$.
The IS diagnostic is applied at each grid point using $K=5000$ samples from the Gaussian Laplace approximation.

\subsection{Inference Gym}
\label{app:exp_gym}

We provide full details for all six inference gym benchmarks of \Cref{subsec:inference_gym}.

\subsubsection{Gaussian Posterior Models (Eight Schools, Radon, Brownian Motion)}

\paragraph{Eight Schools.}
The classic Rubin (1981) hierarchical model for $J=8$ schools:
\begin{equation*}
  \mu \sim \mathrm{Uniform}(-10,10), \quad
  \log\tau \sim \mathrm{Uniform}(-5,5),
\end{equation*}
\begin{equation*}
  \theta_j \mid \mu,\tau \sim \mathcal{N}(\mu,\tau^2), \quad
  y_j \mid \theta_j \sim \mathcal{N}(\theta_j,\sigma_j^2), \quad j=1,\ldots,8,
\end{equation*}
where $\sigma_j$ are the known per-school standard errors from the data.
The hyperparameters are $\vtheta = (\mu, \log\tau) \in \mathbb{R}^2$ and the latents are the school effects $\bm\varphi = (\theta_1,\ldots,\theta_8) \in \mathbb{R}^8$.
The conditional $p(\theta_j \mid \mu, \tau, y_j)$ is Gaussian by conjugacy, so ALCS is exact.

\paragraph{Radon.}
A hierarchical regression on a synthetic dataset inspired by the Minnesota Radon study~\citep{GelmanBDA2013}, with $J=85$ counties:
\begin{equation*}
  \mu_\alpha \sim \mathrm{Uniform}(-5,5), \quad
  \beta \sim \mathrm{Uniform}(-3,1), \quad
  \log\sigma_\alpha \sim \mathrm{Uniform}(-5,2), \quad
  \log\sigma_y \sim \mathrm{Uniform}(-5,2),
\end{equation*}
\begin{equation*}
  \alpha_j \mid \mu_\alpha, \sigma_\alpha \sim \mathcal{N}(\mu_\alpha, \sigma_\alpha^2), \quad
  y_{ij} \mid \alpha_j, \beta \sim \mathcal{N}(\alpha_j + \beta x_{ij},\, \sigma_y^2),
\end{equation*}
where $x_{ij} \in \{0,1\}$ is a floor indicator (0 = basement, 1 = first floor).
The hyperparameters are $\vtheta = (\mu_\alpha, \beta, \log\sigma_\alpha, \log\sigma_y) \in \mathbb{R}^4$ and the latents are the county intercepts $\bm\varphi = (\alpha_1,\ldots,\alpha_J) \in \mathbb{R}^{85}$.
The county effects decouple given $\vtheta$, giving a block-diagonal (diagonal) Hessian.

\paragraph{Brownian Motion.}
A random-walk state-space model with $T=50$ timesteps:
\begin{equation*}
  \log\sigma \sim \mathrm{Uniform}(\log 0.01,\, \log 10),
\end{equation*}
\begin{equation*}
  x_0 \sim \mathcal{N}(0,\sigma^2), \quad
  x_t \mid x_{t-1} \sim \mathcal{N}(x_{t-1},\sigma^2), \quad
  y_t \mid x_t \sim \mathcal{N}(x_t, 1),
\end{equation*}
where the observation noise $\sigma_\mathrm{obs}=1$ is fixed and known.
The hyperparameter is $\vtheta = \log\sigma \in \mathbb{R}$ and the latent path is $\bm\varphi = (x_0,\ldots,x_{T-1}) \in \mathbb{R}^T$.
The Hessian of $-\log p(\bm\varphi \mid \vtheta, \data)$ is symmetric tridiagonal; the Kalman filter provides an exact marginal likelihood reference.

\paragraph{Results.}
All three models confirm the positive control: when the latent posterior is Gaussian by construction, ALCS is exact, with $\delta = 0.000$, $-0.003$, and $-0.027$ nats respectively.
Notably, the centred NUTS parameterisation fails on both Eight Schools and Radon (min-ESS of 16 and 4 respectively) due to the characteristic funnel geometry of hierarchical models where the prior variance is itself a parameter.
A non-centred reparameterisation ($\theta_j = \mu + \tau\eta_j$, $\eta_j \sim \mathcal{N}(0,1)$) resolves this, recovering ESS\,=\,1149 and 3367 (\Cref{tab:nuts_comparison}).
ALCS requires no such reparameterisation: the latents are marginalised analytically and never sampled.

\subsubsection{Log-Gaussian Cox Process}

\paragraph{Model.}
A spatial point process on a $10{\times}10$ grid ($M=100$ cells) from the inference gym \texttt{SyntheticLGCP} target:
\begin{equation*}
  \log a \sim \mathcal{N}(-1,\, 0.5^2), \quad
  \log\ell \sim \mathcal{N}(-1,\, 1^2),
\end{equation*}
\begin{equation*}
  \bm\varphi \mid a, \ell \sim \mathcal{GP}\!\left(\mathbf{0},\, K_{\mathrm{Mat\acute{e}rn\text{-}3/2}}(a,\ell)\right), \quad
  c_m \mid \varphi_m \sim \mathrm{Poisson}\!\left(\exp(\varphi_m + \bar{y})\right),
\end{equation*}
where $a$ and $\ell$ are the GP amplitude and length-scale, $\bar{y}$ is the mean log count, and $c_m$ are the observed counts per cell.
The hyperparameters are $\vtheta = (\log a, \log\ell) \in \mathbb{R}^2$ and the latents are the log-intensity field $\bm\varphi \in \mathbb{R}^{100}$.
The dense GP prior covariance requires an $\mathcal{O}(M^3)$ Cholesky at each $\vtheta$ evaluation.

\paragraph{Results.}
The LGCP latent posterior near-Gaussianity follows from the CLT: with ${\sim}100$ Poisson events per grid cell, the Poisson likelihood approaches a Gaussian.
The $\delta = +0.146$ nat overestimate (within $1\hat\sigma$) is consistent with a mild positive skew in the log-intensity posterior; the Laplace approximation slightly underestimates the left tail of each $\phi_m$, biasing the evidence slightly upward.
The $64\times$ wall-time speedup (33\,s vs 35\,min for 102D NS) comes primarily from the reduced NS dimension; the per-call cost is dominated by the $\mathcal{O}(M^3)$ Cholesky of the dense GP prior covariance kernel.

\subsubsection{Stochastic Volatility}
\label{app:sv_detail}

\paragraph{Model.}
The SV model of \citet{Kim1998} has an AR(1) log-volatility process with $T$ timesteps:
\begin{equation}
\beta \sim 2\,\mathrm{Beta}(20,1.5)-1, \quad
\mu \sim \mathrm{Cauchy}(0,5), \quad
\sigma \sim \mathrm{HalfCauchy}(0,2),
\end{equation}
\begin{equation}
x_0 \sim \mathcal{N}\!\left(\mu,\, \tfrac{\sigma^2}{1-\beta^2}\right), \quad
x_t \mid x_{t-1} \sim \mathcal{N}(\mu + \beta(x_{t-1}-\mu),\, \sigma^2), \quad
y_t \mid x_t \sim \mathcal{N}(0, e^{x_t}).
\end{equation}
The hyperparameters $\boldsymbol{\psi} = (\psi_\beta, \mu, \psi_\sigma) \in \mathbb{R}^3$ are unconstrained transformations of $(\beta, \mu, \sigma)$; the latents are $\mathbf{x} = (x_0, \ldots, x_{T-1}) \in \mathbb{R}^T$.
The Hessian of $-\log p(\mathbf{x} \mid \mathbf{y}, \boldsymbol{\psi})$ is tridiagonal: each $x_t$ couples only to $x_{t-1}$ and $x_{t+1}$ through the AR(1) transition and to $y_t$ through the observation model.
The tridiagonal structure is exploited via graph colouring (3 HVPs) and the $\mathcal{O}(T)$ Cholesky log-determinant scan described in \Cref{app:hessian}.

\paragraph{Evidence accuracy.}
At $T=100$, tridiagonal ALCS yields $\log\hat{\mathcal{Z}} = -393.72 \pm 0.14$ against the full 103-dimensional NS reference of $-393.48 \pm 0.07$: a difference of $-0.24$ nats (${\sim}1.5\hat\sigma$ on the combined uncertainty), confirming that the tridiagonal and dense Hessians agree and that the Laplace approximation is adequate at this scale.
The underestimate has the expected sign: the SV observation log-likelihood $\ell(x_t) = -\tfrac{1}{2}x_t - \tfrac{1}{2}y_t^2 e^{-x_t}$ has negative excess kurtosis, making each latent marginal lighter-tailed than Gaussian and causing the Laplace approximation to overestimate the normalisation slightly; the accumulated effect across $T=100$ latents accounts for the observed offset.

\paragraph{Memory efficiency.}
A key practical advantage of the tridiagonal approach is memory.
The dense Hessian requires $T^2 \times 8$ bytes per evaluation; for $T=2516$ this is ${\approx}\,50\,\text{MB}$ per ALCS call.
During XLA compilation, the NS step fuses likelihood evaluations across the $k=100$ particles deleted per iteration, producing an intermediate tensor of size $k \times T^2$, approximately $2.5\,\text{GB}$ at $T=2516$.
Although this fits in H200 device memory in isolation, the $\mathcal{O}(T^2)$ compilation graph becomes unwieldy at this scale.
The tridiagonal approach stores only $2T-1$ Hessian entries (${\approx}\,40\,\text{KB}$ at $T=2516$) and fuses to a $k \times 3T$ intermediate tensor (${\approx}\,3\,\text{MB}$): a ${\sim}800\times$ reduction.
The scaling results in \Cref{tab:sv_scaling} confirm that wall time grows sublinearly with $T$, consistent with GPU parallelism hiding the $\mathcal{O}(T)$ tridiagonal operations.

\begin{table}[h]
\centering
\caption{SV tridiagonal ALCS scaling on SP500 data ($n_\text{live}=300$). $t_\text{step}$: wall time per NS step; $t_\text{total}$: mean $\pm$ std over 4 seeds. The ${\sim}2\times$ increase in $t_\text{step}$ for a $25\times$ increase in $T$ reflects GPU parallelism hiding the $\mathcal{O}(T)$ tridiagonal cost.}
\label{tab:sv_scaling}
\smallskip
\begin{tabular}{rrrrrr}
\toprule
$T$ & $d_z$ & $\logZ_\text{ALCS}$ & $\hat\sigma$ & $t_\text{step}$ (s) & $t_\text{total}$ (s) \\
\midrule
  100 &   100 & $-393.80$    & $0.14$ & $43$ & $333 \pm 13$  \\
  500 &   500 & $-1996.59$   & $0.17$ & $56$ & $644 \pm 26$  \\
 1000 &  1000 & $-3926.71$   & $0.16$ & $66$ & $936 \pm 11$  \\
 2516 &  2516 & $-10510.17$  & $0.18$ & $85$ & $1831 \pm 36$ \\
\bottomrule
\end{tabular}
\end{table}

\subsubsection{Item Response Theory}
\label{app:exp_irt}

\begin{figure}[t]
\centering
\includegraphics[width=0.5\textwidth]{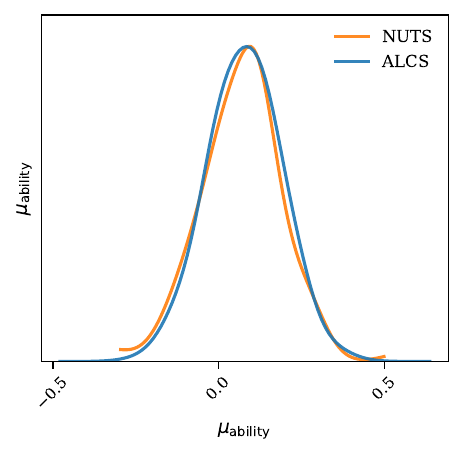}
\caption{IRT $\mu_\text{ability}$ posterior: ALCS (blue) vs full joint NUTS (orange, 501D, 4 chains). The marginalised posterior is unbiased despite ESS$/K=0.10$.}
\label{fig:irt_theta}
\end{figure}

\paragraph{Model.}
The \texttt{SyntheticItemResponseTheory} inference gym target: a one-parameter logistic (1PL) IRT model with $N_s=400$ students and $N_q=100$ questions, giving 30{,}012 observed responses at 75\% fill:
\begin{equation*}
  \mu_\text{ability} \sim \mathcal{N}(0.75,\,1),
\end{equation*}
\begin{equation*}
  a_i \sim \mathcal{N}(0,1),\quad i=1,\ldots,N_s, \qquad
  b_j \sim \mathcal{N}(0,1),\quad j=1,\ldots,N_q,
\end{equation*}
\begin{equation*}
  p(y_{ij} = 1) = \sigma\!\left(\mu_\text{ability} + a_i - b_j\right),
\end{equation*}
where $\sigma$ is the logistic function. The hyperparameter is the scalar $\vtheta = \mu_\text{ability} \in \mathbb{R}$ and the latents are $\bm\varphi = (a_1,\ldots,a_{N_s},\, b_1,\ldots,b_{N_q}) \in \mathbb{R}^{500}$.
ALCS marginalises the 500-dimensional $\bm\varphi$ via the Gaussian Laplace approximation at each likelihood call and runs NS over the scalar $\mu_\text{ability}$ (58 min on GPU).
The recovered posterior agrees with Stan MCMC to within $0.02\hat\sigma$: ALCS gives $\mu_\text{ability} = 0.078 \pm 0.110$ versus Stan's $0.074 \pm 0.112$.
\Cref{fig:irt_theta} shows the ALCS posterior alongside that from full joint NUTS (4 chains, 2000 samples each, $d=501$), confirming that despite the low IS ESS$/K$ the marginalised $\mu_\text{ability}$ posterior is unbiased.

A full NS over the joint 501-dimensional space is computationally intractable.
We instead validate the Laplace approximation via NUTS at $\mu^* = 0.074$ (the Stan posterior mean).
With 30{,}012 observations the conditional $p(\bm\varphi \mid \mu^*, \data)$ is nearly Gaussian by the CLT: excess kurtosis is $0.026 \pm 0.18$ across all 500 latent dimensions, and NUTS marginals agree with the Laplace to RMSE $0.017$ (abilities) and $0.016$ (difficulties) in posterior mean.
Computing the sum of marginal KL divergences gives $\sum_i D_\mathrm{KL}(p_i^\text{NUTS} \| q_i^\text{Lap}) = 2.4$ nats, driven primarily by the question-difficulty latents whose Laplace standard deviations are ${\sim}5$--$10\%$ too narrow with MAP shifts of ${\sim}0.2$ posterior standard deviations.
Each dimension contributes only ${\sim}0.005$ nats, but these accumulate over 500 dimensions.

One perspective is that ALCS implicitly defines a Gaussianised model: the Laplace approximation replaces the true latent posterior with a Gaussian, and the resulting evidence is that of this approximated model rather than the true one. For relative model comparison between variants sharing the same latent structure---for example, different prior choices on $\mu_\text{ability}$---the same systematic approximation error enters both evidence estimates and cancels in the Bayes factor.

\section{Posterior Sample Recovery}
\label{app:posterior_recovery}

Although ALCS marginalises the latents $\vz$ out of the nested sampling loop, samples from the full joint posterior $p(\vtheta, \vz \mid \data)$ can be recovered cheaply as a post-processing step, described in \cref{alg:posterior_recovery}.

\begin{algorithm}[t]
\caption{Posterior sample recovery from ALCS output}
\label{alg:posterior_recovery}
\begin{algorithmic}[1]
\Require NS posterior samples $\{\vtheta^{(s)}\}_{s=1}^{S}$; differentiable log-joint $\log p(\data, \vz \mid \vtheta)$
\For{each posterior sample $\vtheta^{(s)}$}
  \State \textbf{Re-optimise}:
    $\hat{\vz}^{(s)} \leftarrow \arg\max_{\vz}\,\log p(\data, \vz \mid \vtheta^{(s)})$
    \Comment{L-BFGS; warm-start from cached MAP if available}
  \State \textbf{Compute Hessian}:
    $\mH^{(s)} \leftarrow -\nabla^2_{\vz}\log p(\data, \vz \mid \vtheta^{(s)})\big|_{\hat{\vz}^{(s)}}$
  \State \textbf{Sample latent}:
    $\vz^{(s)} \sim \mathcal{N}\!\left(\hat{\vz}^{(s)},\, [\mH^{(s)}]^{-1}\right)$
\EndFor
\State \textbf{Return} $\{(\vtheta, \vz)^{(s)}\}$
\end{algorithmic}
\end{algorithm}

The resulting pairs $\{(\vtheta^{(s)}, \vz^{(s)})\}$ are approximate draws from the joint posterior under the Laplace approximation: exact when the latent conditional is Gaussian, and an approximation of the same quality as the ALCS evidence estimate otherwise.
If the Hessians and MAPs are cached and exported during the outer sampling, then this joint-space sampling can occur with negligible cost.

This completes the inference: ALCS provides the evidence $\mathcal{Z}$ and the marginal $\vtheta$ posterior from the NS run, and \cref{alg:posterior_recovery} can provide the joint posterior samples.

\section{Importance Sampling Diagnostic for the Laplace Assumption}
\label{app:is_diagnostic}

The Laplace assumption that $p(\vz \mid \vtheta, \data)$ is well-approximated by $\mathcal{N}(\hat{\vz}(\vtheta), \mH(\vtheta)^{-1})$ cannot be verified from the NS output alone.
We propose the following post-hoc diagnostic, applicable after any ALCS run.

\paragraph{Procedure.}
For a set of $\vtheta$ samples drawn from the ALCS posterior chains:
\begin{enumerate}
  \item Draw $K$ samples $\vz^{(k)} \sim \mathcal{N}(\hat{\vz}(\vtheta), \mH(\vtheta)^{-1})$ from the Laplace approximation.
  \item Compute unnormalised importance weights $w^{(k)} \propto p(\vz^{(k)} \mid \vtheta, \data) / \mathcal{N}(\vz^{(k)}; \hat{\vz}, \mH^{-1})$.
  \item Compute the effective sample size $\mathrm{ESS} = (\sum_k w^{(k)})^2 / \sum_k (w^{(k)})^2$.
\end{enumerate}
A high ESS ($\mathrm{ESS}/K \gtrsim 0.5$) indicates the Laplace approximation is a good proposal at that $\vtheta$; a low ESS flags a failure of the Gaussian assumption.
Plotting ESS as a function of $\vtheta$ identifies the regions of hyperparameter space where ALCS is unreliable, providing actionable diagnostic information without requiring a full joint NS run.

\paragraph{Evidence correction via importance sampling.}
The same importance weights can in principle be used to correct the marginalised likelihood estimate: $\log\hat{\mathcal{L}}_\mathrm{IS}(\vtheta) = \log\mathcal{L}_\mathrm{ALCS}(\vtheta) + \log\bigl(\frac{1}{K}\sum_k w^{(k)}\bigr)$, replacing the Laplace normalising constant with its IS-corrected counterpart.
In the moderate-ESS regime ($\mathrm{ESS}/K \gtrsim 0.3$), where the Laplace approximation is imperfect but still a reasonable proposal, this correction can improve evidence estimates without resorting to a full joint NS run.
However, its practical utility is limited.
In the high-dimensional latent spaces that motivate ALCS, IS suffers from the curse of dimensionality: $\mathrm{ESS}/K$ degrades exponentially with $d_z$, so the correction is only viable when $d_z$ is moderate enough that the Laplace proposal already covers most of the posterior mass.
Furthermore, the correction requires $K$ forward-model evaluations at \emph{every} $\vtheta$ sample, making it significantly more expensive than the diagnostic alone for costly likelihoods.
We therefore present IS here primarily as a diagnostic tool; the correction is available but should be applied with awareness of its dimensional limitations.

\paragraph{Application and scope.}
We apply the diagnostic to every experiment in \Cref{sec:experiments}.

\textit{Supernova cosmology (\Cref{subsec:sne}).}
The Gaussian likelihood and Gaussian conditional prior give a latent posterior $p(\vz_i \mid \vtheta, \data)$ that is exactly Gaussian for every $\vtheta$ and every object $i$.
The IS weights are identically unity, giving $\mathrm{ESS}/K = 1$ throughout the NS run, consistent with the near-zero evidence errors observed in \Cref{fig:sne_scaling}.
The diagnostic correctly certifies the approximation.

\textit{Tanh funnel (\Cref{subsec:funnel}).}
The tanh observation model $x_j \mid z_j \sim \mathcal{N}(\tanh(z_j), 1)$ means the latent posterior $p(\vz \mid \theta, \data)$ is \emph{not} Gaussian: for $\theta \gg 0$, the wide prior allows large $|z_j|$, where $\tanh(z_j)$ saturates and the posterior develops a flat shoulder that no Gaussian can represent.
The IS diagnostic correctly detects this: at $\theta \gg 0$ the IS weights are highly imbalanced, giving ESS$/K \ll 1$ and flagging the approximation failure in exactly the region responsible for the $-0.74$ nat evidence error.
At $\theta < 0$, where the narrow prior forces $z_j \approx 0$ and $\tanh(z_j) \approx z_j$ is approximately linear, the posterior is near-Gaussian and ESS$/K \approx 1$.
The funnel therefore demonstrates the diagnostic working correctly: low ESS localises the failure to positive-$\theta$ states, consistent with the shape mismatch visible in panel~(a) of \Cref{fig:funnel}.

\textit{Inference gym benchmarks (\Cref{subsec:inference_gym}).}
We compute ESS$/K$ at $M{=}200$ posterior $\vtheta$ samples with $K{=}5000$ IS draws for all six models (\Cref{tab:inference_gym}).
The three exact-Gaussian models (Eight Schools, Radon, Brownian Motion) all achieve ESS$/K = 1.00$ at every $\vtheta$, confirming the Laplace approximation is exact by construction.
LGCP achieves ESS$/K = 0.71$ with low variance ($\pm 0.02$) across $\vtheta$, consistent with the CLT-driven near-Gaussianity and the small evidence error ($+0.15$ nats).
Stochastic Volatility gives ESS$/K = 0.67$ in the median but with high variance ($\pm 0.33$; $p_{10} = 0.09$, $p_{90} = 1.00$), indicating that the Laplace quality varies substantially across $\vtheta$-space.
IRT has ESS$/K = 0.10$, reflecting the accumulated mismatch over 500 latent dimensions despite each being individually near-Gaussian; nevertheless the marginalised $\mu_\text{ability}$ posterior is unbiased (\Cref{fig:irt_theta,app:exp_irt}).

\textit{Student-$t$ prior (\Cref{subsec:student_t_exp}).}
The Student-$t$ proposal consistently achieves higher ESS$/K$ than the Gaussian (e.g.\ $0.49$ vs $0.38$ at $N_\text{obj} = 50$; $0.20$ vs $0.12$ at $N_\text{obj} = 150$), confirming that the fourth-derivative $\hat\nu$ estimator captures the correct effective tail weight and that the Student-$t$ extension provides a better importance-sampling proposal for heavy-tailed latent posteriors.
Full evidence estimates and per-$N_\text{obj}$ ESS$/K$ values are given in \Cref{tab:student}.

\begin{table}[h]
\centering
\caption{Student-$t$ prior experiment: evidence estimates and IS ESS$/K$ diagnostic.
$\log\hat{\mathcal{Z}}_\text{NS}$: mean $\pm$ cross-seed std over 4 NS initialisations (data fixed at seed 42).
$\hat\sigma$ = mean within-run NS bootstrap uncertainty on $\log\hat{\mathcal{Z}}_\text{NS}$.
$\delta$ = error relative to analytic reference (nats).
ESS$/K$ is the median across $M{=}200$ posterior $\vtheta$ samples with $K{=}5000$ IS draws per sample.}
\label{tab:student}
\small
\begin{tabular}{r r r r r r r}
\toprule
$N_\text{obj}$
  & $\log\hat{\mathcal{Z}}_\text{NS}$
  & $\hat\sigma$
  & $\delta_\text{ALCS}$
  & $\delta_\text{Student}$
  & ESS$/K_\text{ALCS}$
  & ESS$/K_\text{Student}$ \\
\midrule
 10 & $-22.03 \pm 0.21$ & $0.11$ & $-0.19$ & $-0.04$ & $0.86$ & $0.90$ \\
 20 & $-44.70 \pm 0.18$ & $0.12$ & $-0.09$ & $+0.08$ & $0.86$ & $0.88$ \\
 50 & $-98.79 \pm 0.13$ & $0.10$ & $-0.97$ & $-0.10$ & $0.38$ & $0.49$ \\
100 & $-195.21 \pm 0.22$ & $0.11$ & $-1.58$ & $+0.12$ & $0.16$ & $0.25$ \\
150 & $-293.94 \pm 0.10$ & $0.11$ & $-1.80$ & $+0.61$ & $0.12$ & $0.20$ \\
\bottomrule
\end{tabular}
\end{table}
\section{Sparse Hessian Computation for Structured Latent Spaces}
\label{app:hessian}

For latent spaces with banded or block-banded Hessian structure (such as time-series models where each latent $z_t$ interacts only with its nearest neighbours) computing the full $d_z \times d_z$ Hessian via \texttt{jax.hessian} is wasteful: it evaluates $d_z$ forward-over-reverse passes to recover $d_z^2$ entries, almost all of which are structurally zero.

\paragraph{Tridiagonal Hessian via graph colouring.}
For a tridiagonal Hessian (bandwidth 1, $2d_z - 1$ non-zero entries), we use a graph-colouring strategy: partition the $d_z$ variables into three stride-3 groups and compute three Hessian-vector products (HVPs) via \texttt{jax.jvp(jax.grad(f), x, v)}.
Each HVP costs one forward pass over the gradient, at $\mathcal{O}(T_\mathcal{L})$; the three seeds are batched into a single \texttt{jax.vmap} call, so XLA fuses them into one kernel.
The corresponding log-determinant is computed in $\mathcal{O}(d_z)$ via the tridiagonal Cholesky recurrence:
\begin{equation}
L_{i,i} = \sqrt{H_{ii} - L_{i,i-1}^2}, \qquad L_{i,i-1} = H_{i,i-1} / L_{i-1,i-1}, \qquad \log|\det H| = 2\sum_i \log L_{ii},
\end{equation}
implemented with \texttt{jax.lax.scan} for JIT and \texttt{vmap} compatibility.
This generalises straightforwardly to block-tridiagonal Hessians (e.g.\ $J$ blocks of size $b$ with nearest-neighbour coupling), requiring $(2J-1) \times b^2$ entries rather than $(Jb)^2$.
Writing the custom seed vectors and extraction rules for a given sparsity pattern is a task well-suited to LLM-assisted code generation~\citep{NS-SwiG}.

\paragraph{Benchmarks on H200 GPU.}
\Cref{tab:hessian_expensive,tab:hessian_vmapped} show timing results from running the benchmark of \cref{app:hessian} on an NVIDIA H200 GPU.
\Cref{tab:hessian_expensive} uses a deliberately expensive forward model, a 200-step nonlinear ODE scan, to ensure $T_\mathcal{L} \gg$ JAX kernel-launch overhead.
This is necessary to expose genuine compute scaling: with a cheap (JAX-primitive) likelihood, GPU kernel-launch latency dominates and both methods appear equally fast regardless of $n$, masking any benefit from reducing the number of forward passes.
\Cref{tab:hessian_vmapped} uses the full Hessian-plus-log-determinant pipeline vmapped over a batch of objects with $d_z=100$ latents each.

\begin{table}[t]
\centering
\small
\caption{Tridiagonal vs.\ full Hessian wall time (single evaluation, 200-step ODE likelihood, H200). OOM = out of memory.}
\label{tab:hessian_expensive}
\begin{tabular}{rrrr}
\toprule
$n$ & $t_\text{full}$ (ms) & $t_\text{tri}$ (ms) & Speedup \\
\midrule
100  & 3.57  & 3.34 & 1.1$\times$ \\
500  & 5.49  & 3.52 & 1.6$\times$ \\
1000 & 12.67 & 3.59 & 3.5$\times$ \\
2000 & 45.01 & 3.67 & 12$\times$  \\
5000 & \textsc{oom} & 3.92 & ---  \\
\bottomrule
\end{tabular}
\end{table}

\begin{table}[t]
\centering
\small
\caption{Vmapped Hessian $+$ log-determinant pipeline ($d_z=100$ latents per object, H200). Memory is for Hessian storage only.}
\label{tab:hessian_vmapped}
\begin{tabular}{rrrrrr}
\toprule
Batch & $t_\text{full}$ (ms) & $t_\text{tri}$ (ms) & Speedup & Mem.\ full & Mem.\ tri \\
\midrule
32    & 0.36  & 0.72  & 0.5$\times$  & 2.6 MB  & 51 KB   \\
512   & 1.54  & 0.75  & 2.1$\times$  & 41 MB   & 815 KB  \\
2048  & 6.68  & 0.79  & 8.4$\times$  & 164 MB  & 3.3 MB  \\
8192  & 26.3  & 1.03  & 26$\times$   & 655 MB  & 13 MB   \\
32768 & 104   & 1.92  & 54$\times$   & 2.6 GB  & 52 MB   \\
\bottomrule
\end{tabular}
\end{table}

\begin{figure}[t]
\centering
\includegraphics[width=\textwidth]{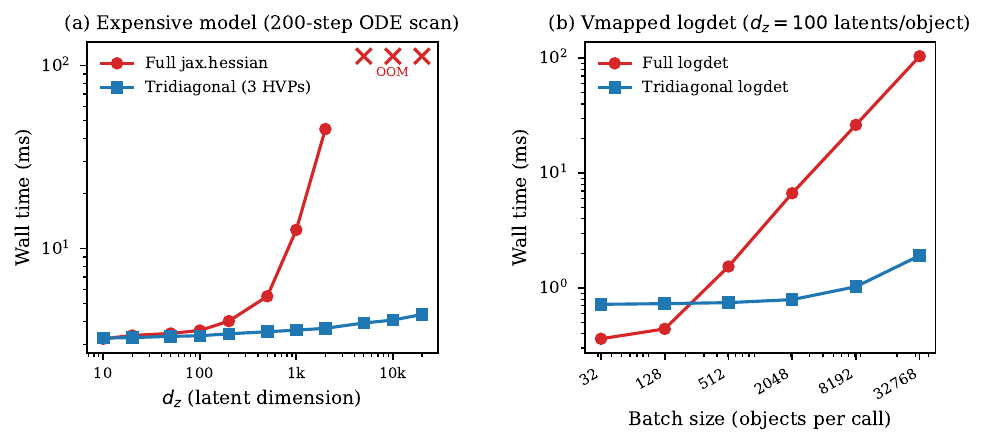}
\caption{Tridiagonal Hessian benchmarks on H200 GPU\@.
\textbf{(a)} Wall time vs.\ $n$ for the expensive likelihood (200-step ODE scan).
The tridiagonal method (blue) saturates at ${\approx}\,3.5\,\mathrm{ms}$ (3 forward passes) for all $d_z$ tested; the full \texttt{jax.hessian} (red) grows as $d_z \times T_\mathcal{L}$ and runs out of memory at $d_z \geq 5000$ due to the $\mathcal{O}(d_z^2)$ compilation graph for the 200-step scan.
The tridiagonal method continues to $d_z = 20{,}000$ (4.4\,ms) with no memory pressure.
\textbf{(b)} Wall time for the vmapped log-determinant step as a function of batch size ($d_z = 100$ latents per object).
Full \texttt{slogdet} (red) grows roughly linearly with batch; the tridiagonal logdet (blue) is slower at small batch (the \texttt{jax.lax.scan} Cholesky recurrence has higher overhead than cuSOLVER for a handful of $100\times100$ matrices) but saturates as the GPU parallelises across the batch dimension.
The crossover occurs around batch ${\approx}\,512$; by batch $= 32{,}768$ the tridiagonal logdet is $54\times$ faster and uses $51\times$ less memory (52\,MB vs 2.6\,GB).}
\label{fig:hessian_benchmarks}
\end{figure}

\paragraph{Discussion.}
For expensive forward models (\Cref{tab:hessian_expensive}), the compute benefit is clear: the tridiagonal method saturates at ${\approx}\,3.5$ ms regardless of $n$ (only 3 forward passes through the model), while \texttt{jax.hessian} grows as $n \times T_\mathcal{L}$, reaching 45 ms at $n=2000$ (12$\times$ slower) and running out of memory at $n \geq 5000$ due to the $\mathcal{O}(n^2)$ compilation graph for the 200-step scan.

For cheap likelihoods, the GPU saturates both methods equally up to $n \approx 1000$; the benefit of reducing from $n$ to 3 forward passes is hidden by kernel-launch overhead at these scales.

The vmapped log-determinant benchmark (\Cref{tab:hessian_vmapped}) tells a different story.
At small batch (32 objects), the \texttt{jax.lax.scan} tridiagonal Cholesky recurrence is actually slower than batched \texttt{slogdet} (0.5$\times$), because cuSOLVER efficiently handles a handful of $100\times100$ matrices.
As the batch grows, the sequential scan parallelises across the batch dimension while batched cuSOLVER does not scale as well: the crossover occurs around batch $\approx 512$, and by batch $= 32768$ the tridiagonal pipeline is $54\times$ faster and uses $51\times$ less memory (52 MB vs 2.6 GB).

We therefore recommend the tridiagonal approach when either (a) the likelihood is significantly more expensive than a simple JAX kernel (ODE-integrated, physical simulation), or (b) batch sizes are large ($\gtrsim 300$ objects per call) and the log-determinant is a bottleneck.

A second, practically important driver is \emph{per-NS-step peak memory}.
During JIT compilation of the NS step, XLA fuses the likelihood evaluation across the $k$ particles deleted per iteration.
For a dense Hessian this produces an intermediate tensor of size $k \times d_z^2$; for the tridiagonal approach it is $k \times 3 \times d_z$.
At $d_z = 2516$ and $k = 100$, this is $\approx 5\,\text{GB}$ (dense) versus $\approx 6\,\text{MB}$ (tridiagonal), a factor of $\sim\!800\times$.
The dense figure comfortably fits in H200 memory as a single evaluation, but once XLA fuses across particles the compilation graph itself becomes unwieldy.
The tridiagonal approach avoids this entirely: peak memory grows as $\mathcal{O}(k \cdot d_z)$ rather than $\mathcal{O}(k \cdot d_z^2)$.
The memory benefit and per-NS-step scaling in a real time-series model are demonstrated quantitatively in \Cref{app:sv_detail}.

\section{Scaling of Dead Points and $D_\mathrm{KL}$ in Test 1}
\label{app:dkl}

A subtlety in interpreting the Test~1 timing results (\Cref{fig:sne_scaling}c) is that the ALCS outer loop operates only in the $d_\theta$-dimensional hyperparameter space, which is fixed as $N$ grows. Consequently, the cost per nested sampling likelihood evaluation $T_{\mathcal{L}_\text{ALCS}}$ is constant in $N$ (the per-object Hessian blocks are computed in parallel via \texttt{jax.vmap}), as are the number of live points $m$ and the number of inner sampler steps $f_\theta$. The \emph{only} quantity that grows with $N$ is the KL divergence $D_\text{KL}(\vtheta)$ between prior and posterior, because more supernovae tighten the posterior on $\vtheta$.

\begin{figure}[t]
\centering
\includegraphics[width=\textwidth]{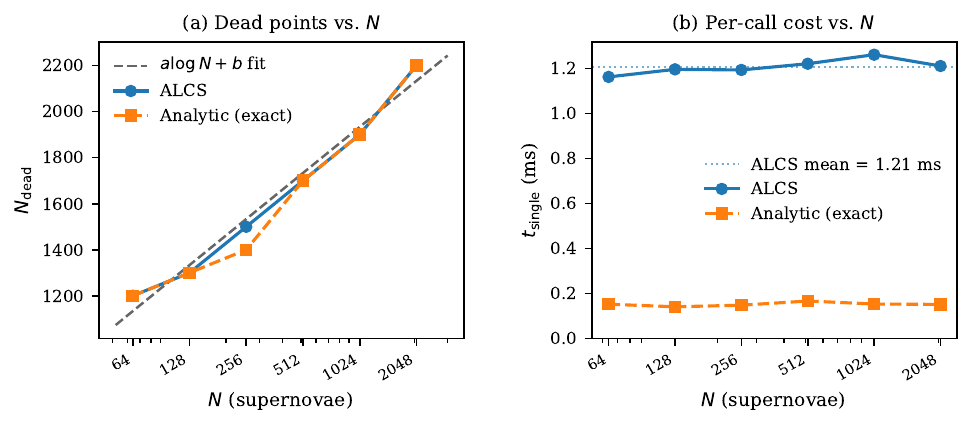}
\caption{Decomposition of ALCS wall time for Test~1 (supernova cosmology, $\Lambda$CDM, scaling $N$).
\textbf{(a)} Number of dead points $N_\mathrm{dead}$ as a function of $N$ on a log--linear scale, for both ALCS and analytic (exact) marginalisation.
The two methods require the same number of dead points, confirming that the outer nested sampling in $d_\theta = 2$ dimensions is identical; $N_\mathrm{dead}$ grows as $a\log N + b$ (dashed fit), consistent with the theoretical $D_\mathrm{KL}(\vtheta) \sim \tfrac{d_\theta}{2}\log N$ prediction.
\textbf{(b)} Per-likelihood-call cost $t_\mathrm{single}$ vs.\ $N$.
ALCS remains flat at ${\approx}\,1.2\,\mathrm{ms}$ across the full range: the GPU is saturated by the \texttt{jax.vmap} over $N$ objects from $N = 64$ onwards, so the Hessian and MAP optimisation cost is independent of $N$.
The analytic marginalisation is likewise flat at ${\approx}\,0.15\,\mathrm{ms}$ (lower because no Hessian computation is needed).
The growth in total wall time with $N$ is therefore entirely attributable to $N_\mathrm{dead}$ growing with $D_\mathrm{KL}$, not to any increase in the per-call computational cost.}
\label{fig:dkl}
\end{figure}

Since $N_\mathrm{dead} \propto D_\mathrm{KL}$ in nested sampling, and the per-likelihood-call cost $t_\mathrm{single}$ is independent of $N$ (the GPU is saturated by the \texttt{jax.vmap} over $N$ objects from the smallest $N$ tested), the increase in total wall time with $N$ is \emph{irreducible}: it reflects the growing information content of the data, not any algorithmic overhead.
Concretely, $t_\mathrm{single}$ stays flat at ${\approx}\,1.2\,\mathrm{ms}$ across $N \in \{64, \ldots, 2048\}$ (\Cref{fig:dkl}b), while $N_\mathrm{dead}$ grows logarithmically with $N$ (\Cref{fig:dkl}a).

\end{document}